\documentclass{article}

\PassOptionsToPackage{numbers, compress}{natbib}

\usepackage[final]{neurips_2024}

\usepackage[utf8]{inputenc} 
\usepackage[T1]{fontenc}    
\usepackage[colorlinks=true,citecolor=red]{hyperref} 
\usepackage{url}            
\usepackage{booktabs}       
\usepackage{amsfonts}       
\usepackage{nicefrac}       
\usepackage{microtype}      
\usepackage{xcolor}         

\usepackage{color}
\usepackage{arydshln}
\usepackage{graphicx}
\usepackage{amsmath,bm}
\usepackage{amssymb}
\usepackage{array}
\usepackage{comment}
\usepackage{booktabs}
\usepackage{subfigure}
\usepackage{caption}
\usepackage{colortbl} 
\usepackage{multirow}
\usepackage{makecell}
\usepackage{wrapfig}

\usepackage{subcaption}
\usepackage{lineno}
\usepackage{tikz}
\usepackage[ruled,vlined]{algorithm2e}
\usepackage[normalem]{ulem}
\usepackage{arydshln} 

\usepackage{pifont}

\usepackage{tabularray}
\usepackage{paralist}
\usepackage{makecell}
\usepackage{colortbl}
\usepackage{etoolbox}

\usepackage{tcolorbox}
\usepackage{adjustbox}
\usepackage{rotating}
\tcbuselibrary{breakable}

\definecolor{myblue}{RGB}{52,218,247}
\definecolor{myred}{RGB}{255,90,90}
\definecolor{mypink}{RGB}{239,43,159}
\definecolor{myupdate}{RGB}{254,243,222}
\definecolor{myfrozen}{RGB}{237,255,255}
\definecolor{ired}{RGB}{229,72,72}
\definecolor{igreen}{RGB}{80,219,144}
\definecolor{nmblue}{RGB}{216,234,247}
\definecolor{lightgreen}{RGB}{196,250,222}

\definecolor{linkred}{RGB}{255,0,49}

\definecolor{textred}{RGB}{255,242,234}
\definecolor{imageblue}{RGB}{224,250,255}
\definecolor{videogreen}{RGB}{214,250,232}

\newcommand{\mlogo}{\raisebox{-5pt}{\includegraphics[width=1.5em]{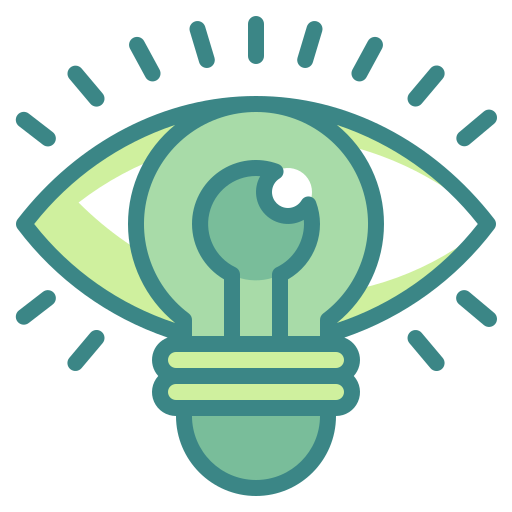}}\xspace}

\newcommand{\textlogo}{\raisebox{-3pt}{\includegraphics[width=1.3em]{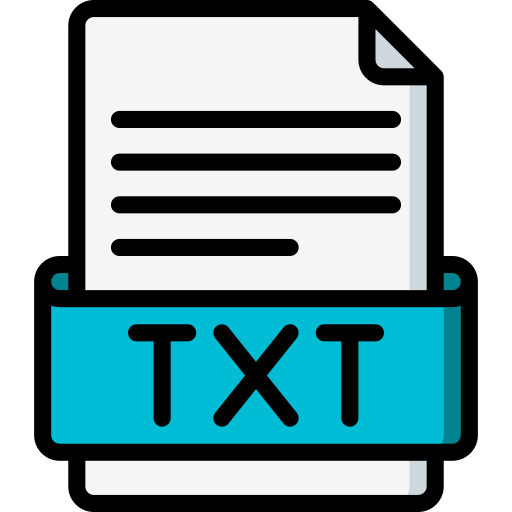}}\xspace\xspace}
\newcommand{\imagelogo}{\raisebox{-4pt}{\includegraphics[width=1.5em]{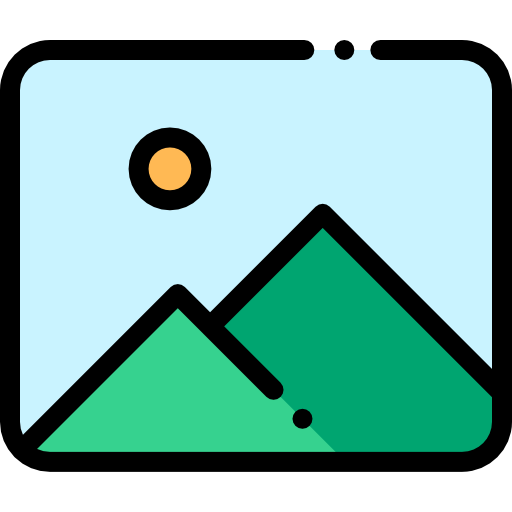}}\xspace\xspace}
\newcommand{\imagereflogo}{\raisebox{-4pt}{\includegraphics[width=1.5em]{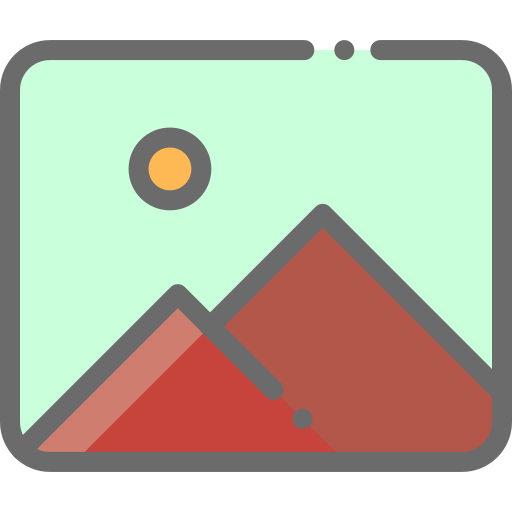}}\xspace\xspace}
\newcommand{\imagereflogosmall}{\raisebox{-3pt}{\includegraphics[width=1.2em]{figures/image-ref.png}}\xspace}
\newcommand{\videologo}{\raisebox{-4pt}{\includegraphics[width=1.5em]{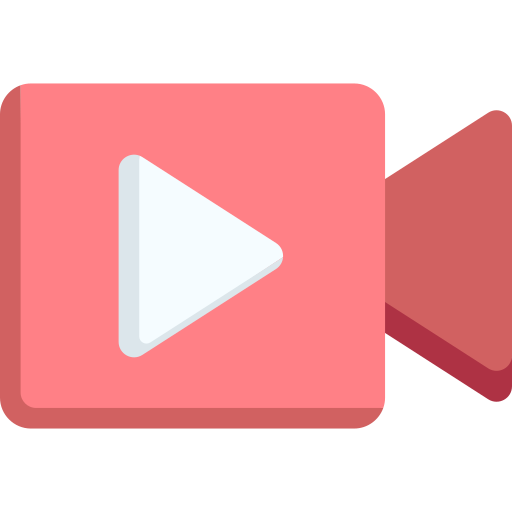}}\xspace\xspace}
\newcommand{\sketchlogo}{\raisebox{-3pt}{\includegraphics[width=1.3em]{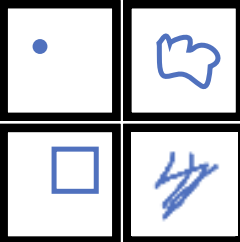}}\xspace\xspace}

\newcommand{\circlenum}[1]{%
    \resizebox{!}{0.8em}{%
        \tikz[baseline=(char.base)]{
            \node[shape=circle, fill=black, inner sep=0.8pt, text=white] (char) {#1};
        }%
    }%
}

\title{\mlogo\textsc{Vitron}: A Unified Pixel-level Vision LLM for\\ Understanding, Generating, Segmenting, Editing}

\author{%
Hao Fei$^{1,2}$ \quad Shengqiong Wu$^{1,2}$ \quad  Hanwang Zhang$^{1,3}$ \quad  Tat-Seng Chua$^{2}$ \quad Shuicheng Yan$^{1,}$\thanks{Shuicheng Yan is the corresponding author. This work was performed when Hao Fei was an Associate Member, and Shengqiong Wu was an Intern at Skywork AI.} \\
$^{1,*}$Skywork AI, Singapore \, $^2$ National University of Singapore \, $^3$ Nanyang Technological University\\
\texttt{haofei37@nus.edu.sg \quad swu@u.nus.edu \quad hanwangzhang@ntu.edu.sg}\\
\texttt{dcscts@nus.edu.sg \qquad shuicheng.yan@kunlun-inc.com}
}

\begin{document}

\maketitle

\vspace{-8mm}
\begin{center}
    \large{\color{mypink} \textbf{Project Homepage:} \url{https://vitron-llm.github.io/}}
\end{center}

\begin{figure}[!h]
  \vspace{-2mm}
\centering
\includegraphics[width=1.0\columnwidth]{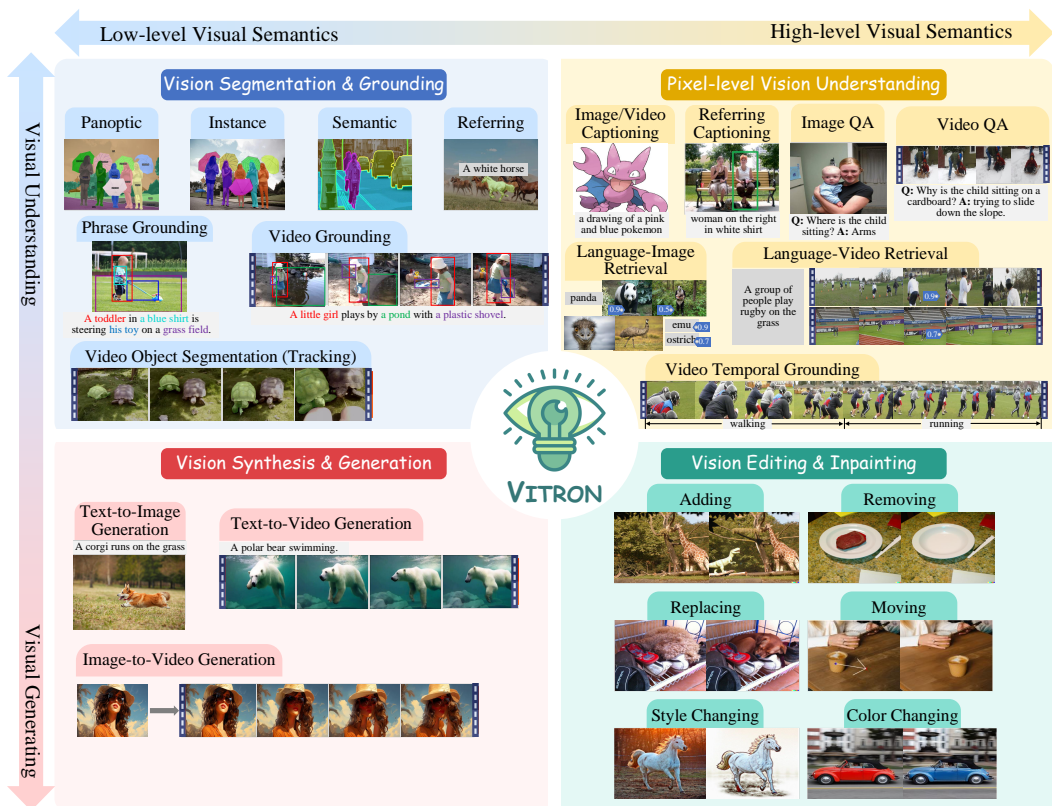}
\caption{
\textsc{Vitron} supports four main task clusters of visions, spanning visual comprehension to visual generation, from low level to high level.
}
\label{fig:tasks}
\end{figure}

\vspace{2mm}
\begin{abstract}
Recent developments of vision large language models (LLMs) have seen remarkable progress, yet still encounter challenges towards multimodal generalists, such as coarse-grained instance-level understanding, lack of unified support for both images and videos, and insufficient coverage across various vision tasks. 
In this paper, we present \textbf{\textsc{Vitron}}, a universal pixel-level vision LLM designed for comprehensive \emph{understanding}, \emph{generating}, \emph{segmenting}, and \emph{editing} of both static images and dynamic videos.
Building on top of an LLM backbone, \textsc{Vitron} incorporates encoders for images, videos, and pixel-level regional visuals within its frontend modules, while employing state-of-the-art visual specialists as its backend, via which \textsc{Vitron} supports a spectrum of vision end tasks, spanning visual comprehension to visual generation, from low level to high level.
To ensure an effective and precise message passing from LLM to backend modules for function invocation, we propose a novel hybrid method by simultaneously integrating discrete textual instructions and continuous signal embeddings.
Further, we design various pixel-level spatiotemporal vision-language alignment learning for \textsc{Vitron} to reach the best fine-grained visual capability.
Finally, a cross-task synergy module is advised to learn to maximize the task-invariant fine-grained visual features, enhancing the synergy between different visual tasks.
Demonstrated over 12 visual tasks and evaluated across 22 datasets, \textsc{Vitron} showcases its extensive capabilities in the four main vision task clusters.
Overall, this work illuminates the great potential of developing a more unified multimodal generalist.
\end{abstract}

\vspace{-12pt}

\section{Introduction}
\label{Introduction}

\vspace{-3mm}

Recently, the field of multimodal large language models (MLLMs) has witnessed rapid and flourishing development across multiple communities. 
Extensive research efforts have been directed towards augmenting powerful, purely language-based LLMs with modules capable of visual perception, thereby extending their applicability to MLLMs \cite{AlayracDLMBHLMM22,0008LSH23,abs-2304-08485,zhang2024vpgtrans,gpt4,wu2024controlmllm,fei2024multimodal}. 
MLLMs, such as BLIP-2 \cite{0008LSH23}, LLaVA \cite{abs-2304-08485}, MiniGPT-4 \cite{abs-2304-10592} and GPT-4V \cite{yang2023dawn} etc., demonstrate a robust and exceptional capability in image understanding, paralleling the deep semantic comprehension of language. 
In the realm of vision, the ability to process and comprehend dynamic videos is equally critical. 
Concurrently, several MLLMs have emerged with a focus on video understanding, e.g., VideoChat \cite{abs-2305-06355} and Video-LLaMA \cite{abs-2306-02858}, demonstrating significant advancements in video comprehension.

Subsequent studies have sought to further expand the capabilities of MLLMs, with efforts bifurcating into two primary dimensions. 
On one hand, there's a deepening of MLLMs' understanding of vision, transitioning from coarse, instance-level comprehension towards a pixel-level, fine-fined understanding of images, thereby achieving visual regional grounding capabilities, as seen in GLaMM \cite{rasheed2023glamm}, PixelLM \cite{ren2023pixellm}, and MiniGPT-v2 \cite{chen2023minigpt}, etc., alongside the counterparts in pixel-grounding video LLMs \cite{munasinghe2023pg}. 
On the other hand, there's an expansion in the breadth of functionalities MLLMs can support within the vision field. 
A portion of the research has already ventured into enabling MLLMs not just to comprehend input vision signals but also to support the generation and output of vision content, with systems like GILL \cite{koh2023generating}, Emu \cite{sun2023generative}, etc., flexibly generating image content, and GPT4Video \cite{wang2023gpt4video} and NExT-GPT \cite{wu2023next} achieving video generation.

We posit that the future trend of vision LLMs necessarily involves the enhancement of their capabilities towards a high degree of unification, i.e., multimodal generalists. 
However, our observations reveal that despite the diversity of existing vision LLMs developed by the community, there is still a clear lack of unification.
\textbf{First}, almost all existing vision LLMs treat images and videos as separate entities, either supporting only images or videos \cite{AlayracDLMBHLMM22,sun2023generative,abs-2304-10592,abs-2306-02858}. 
We argue for a unified vision MLLM framework that concurrently supports both images and videos, acknowledging that vision inherently comprises both static images and dynamic videos - both core components of our world and largely interchangeable in most scenarios. 
\textbf{Second}, the current support for vision functionalities in MLLMs is found wanting, with most models only capable of understanding \cite{abs-2304-08485,abs-2304-10592}, or at most generating images or videos \cite{dong2023dreamllm,wang2023gpt4video}. 
We contend that future MLLMs should embrace a broader spectrum of vision tasks and functionalities, enabling unified support for all vision-related tasks and achieving an ``\emph{one for all}'' capability, which is vital for real-world applications, especially in vision creation that often involves a series of iterative and interactive operations.
For example, users typically start by \emph{generating} images from text, transforming an idea into visual content; and then refining this content through further fine-grained \emph{editing} to add more details; following, proceeding to create dynamic content by \emph{generating} videos from the images; and finally, engaging in several rounds of iterative interaction, such as video \emph{editing}, to enhance and finalize their creation.
\textbf{Last but not the least}, for a generalist integrated with various multimodal functionalities, one key lies in how to ensure that all tasks achieve their best performance as much as possible. 
This includes both that, 1) the instructions from the LLM are precisely conveyed to the downstream decoders, and 2) different tasks do not undermine each other but rather cooperate.

\begin{figure}[!t]
\centering
\includegraphics[width=0.9\columnwidth]{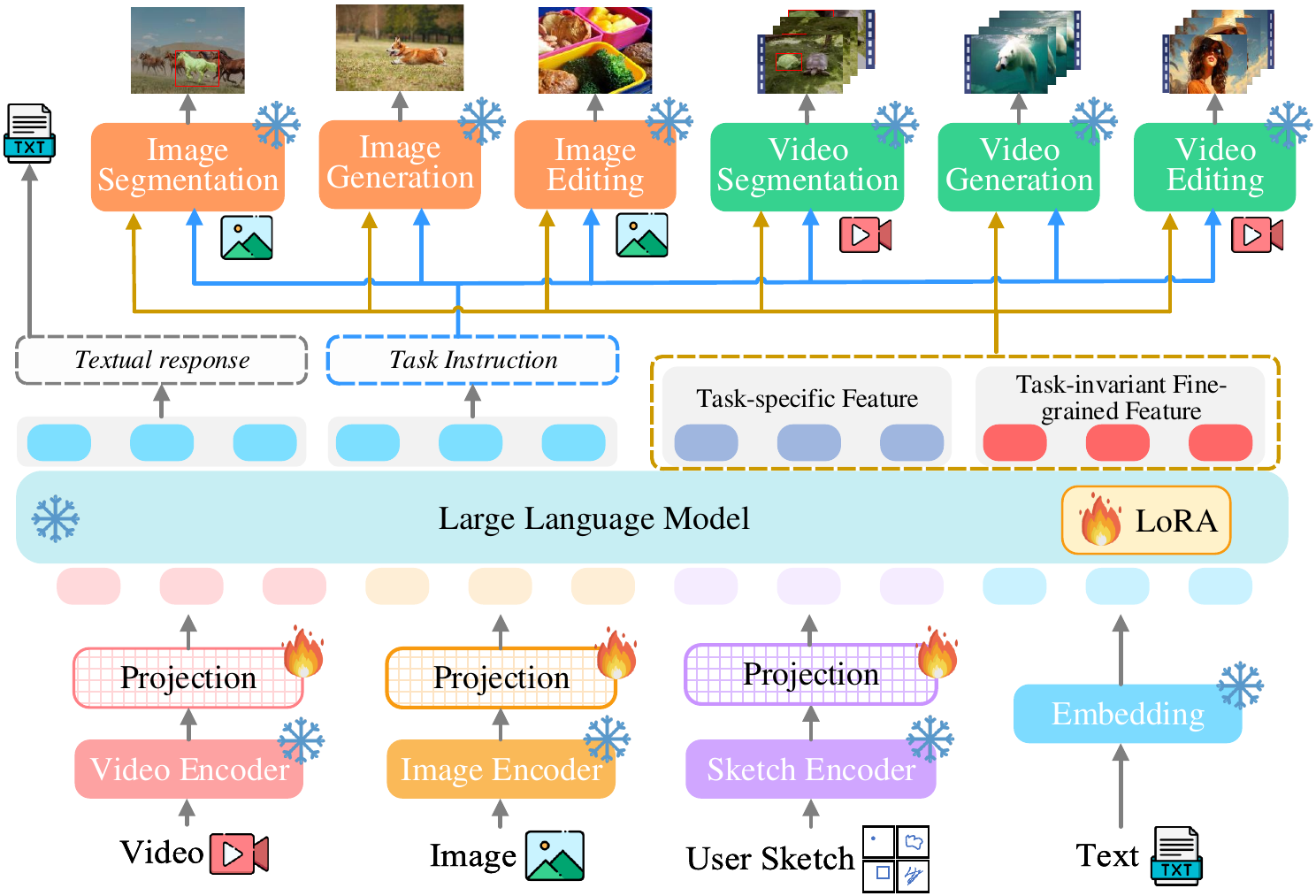}
 \vspace{-2mm}
\caption{
Technical overview of the \textsc{Vitron} framework.
}
  \vspace{-6mm}
\label{fig:framework}
\end{figure}

To address all these gaps, this paper introduces \textbf{\textsc{Vitron}}, a pioneering universal pixel-level vision LLM, as shown in Fig. \ref{fig:framework}. 
First, \textsc{Vitron} leverages a backbone LLM for comprehending, reasoning, decision-making, and multi-round user interactions. 
To perceive both image and video modal signals and support fine-grained user visual inputs, \textsc{Vitron} incorporates encoders for images, videos, and regional box/sketch-specified inputs.
On the backend, several state-of-the-art (SoTA) image and video modules are integrated for decoding and executing a wide range of vision tasks, spanning from lower to higher levels, such as visual understanding (perceiving and reasoning), generating, segmenting (grounding and tracking), editing (inpainting).
To ensure that \textsc{Vitron} precisely conveys the LLM's decisions to various backend decoder modules for function invocation, we propose a novel hybrid method of instruction passing. 
Specifically, we enable the LLM to output not only discrete textual instructions, but also continuous signal feature embeddings passed to the modules.
Finally, to maximize the functionalities of different modules within \textsc{Vitron}, we further devise a synergy module, where we fully maximize the task-persistent fine-grained visual features to be shared among different visual tasks.

The overall training for \textsc{Vitron} aims to equip it with robust and powerful vision understanding and manipulation capabilities. 
We first imbue \textsc{Vitron} basic MLLM skills by carrying out 1) vision-language alignment learning between the frontend encoders and central LLM, also 2) invocation-oriented instruction tuning, and 3) embedding-oriented alignment tuning between LLM and backend modules. 
Going beyond this, we further try to strengthen \textsc{Vitron}'s capacities.
On the one hand, we introduce fine-grained spatiotemporal vision grounding instruction tuning, training LLM on grounding predictions and pixel-aware perception for images and videos, such that \textsc{Vitron} sufficiently gains pixel-level visual perception.
On the other hand, we utilize adversarial training \cite{ganin2016domain,tramer2017ensemble} to decouple \emph{task-specific features} from \emph{task-invariant fine-grained visual features} in signal feature representations, thereby enhancing the synergy between different tasks.

Extensive experiments covering 12 tasks across 22 datasets are performed. 
Leveraging its advanced architecture as a multimodal generalist, \textsc{Vitron} demonstrates proficiency in a comprehensive range of vision tasks. 
Notably, the unified system's performance is on par with or even surpasses singleton state-of-the-art specialists on specific tasks. 
Further analyses reveal the efficacy of each design of the system.
Our overall contributions are summarized as follows.

\circlenum{1} To our knowledge, we for the first time propose a grand unified vision MLLM, \textsc{Vitron}, capable of pixel-level 
understanding, generating, segmenting, editing of both images and videos.
\circlenum{2} We introduce a more effective LLM-to-decode instruction-passing mechanism over both discrete texts and continuous signal embeddings.
\circlenum{3} We propose carrying out various pixel-level vision-language spatiotemporal alignment learning for MLLMs to reach the best fine-grained visual capability.
\circlenum{4} We devise a synergy module to maximize the task-persistent fine-grained visual features shareable among all different visual tasks, via which \textsc{Vitron} surpasses existing SoTA specialists' performance.

\begin{table*}[!t]
\centering
\fontsize{8}{9}\selectfont
\setlength{\tabcolsep}{1.3mm}
\vspace{-2mm}
\begin{tabular}{lccccccc}
\toprule
\multirow{2}{*}{\bf Model }& \multicolumn{2}{c}{\bf Vision Supporting} & \multirow{2}{*}{\makecell[c]{\textbf{Pixel/Regional}\\\textbf{Understanding}}}  & \multirow{2}{*}{\makecell[c]{\textbf{Segmenting/}\\\textbf{Grounding}}} & \multirow{2}{*}{\bf Generating}&  \multirow{2}{*}{\bf Editing } & \multirow{2}{*}{\makecell[c]{\textbf{Cross-task}\\\textbf{Synergy}}} \\ 
\cmidrule(r){2-3} 
&  \textbf{Image} & \textbf{Video} &  &  &  &   \\
\hline

Flamingo \cite{AlayracDLMBHLMM22} & \textcolor{igreen}{\ding{51}} & \textcolor{ired}{\ding{55}} & \textcolor{ired}{\ding{55}} & \textcolor{ired}{\ding{55}} & \textcolor{ired}{\ding{55}} & \textcolor{ired}{\ding{55}}& \textcolor{ired}{\ding{55}} \\
BLIP-2 \cite{0008LSH23} & \textcolor{igreen}{\ding{51}} & \textcolor{ired}{\ding{55}} & \textcolor{ired}{\ding{55}} & \textcolor{ired}{\ding{55}} & \textcolor{ired}{\ding{55}} & \textcolor{ired}{\ding{55}}& \textcolor{ired}{\ding{55}} \\
MiniGPT-4 \cite{abs-2304-10592} & \textcolor{igreen}{\ding{51}} & \textcolor{ired}{\ding{55}} & \textcolor{ired}{\ding{55}} & \textcolor{ired}{\ding{55}} & \textcolor{ired}{\ding{55}} & \textcolor{ired}{\ding{55}} & \textcolor{ired}{\ding{55}}\\
LLaVA \cite{abs-2304-08485} & \textcolor{igreen}{\ding{51}} & \textcolor{ired}{\ding{55}} & \textcolor{ired}{\ding{55}} & \textcolor{ired}{\ding{55}} & \textcolor{ired}{\ding{55}} & \textcolor{ired}{\ding{55}} & \textcolor{ired}{\ding{55}}\\
GILL \cite{koh2023generating} & \textcolor{igreen}{\ding{51}} & \textcolor{ired}{\ding{55}} & \textcolor{ired}{\ding{55}} & \textcolor{ired}{\ding{55}} & \textcolor{igreen}{\ding{51}} & \textcolor{ired}{\ding{55}} & \textcolor{ired}{\ding{55}}\\
Emu \cite{sun2023generative} & \textcolor{igreen}{\ding{51}} & \textcolor{ired}{\ding{55}} & \textcolor{ired}{\ding{55}} & \textcolor{ired}{\ding{55}} & \textcolor{igreen}{\ding{51}} & \textcolor{ired}{\ding{55}} & \textcolor{ired}{\ding{55}}\\
MiniGPT-5 \cite{zheng2023minigpt} & \textcolor{igreen}{\ding{51}} & \textcolor{ired}{\ding{55}} & \textcolor{ired}{\ding{55}} & \textcolor{ired}{\ding{55}} & \textcolor{igreen}{\ding{51}} & \textcolor{ired}{\ding{55}} & \textcolor{ired}{\ding{55}}\\
DreamLLM \cite{dong2023dreamllm} & \textcolor{igreen}{\ding{51}} & \textcolor{ired}{\ding{55}} & \textcolor{ired}{\ding{55}} & \textcolor{ired}{\ding{55}} & \textcolor{igreen}{\ding{51}} & \textcolor{ired}{\ding{55}} & \textcolor{igreen}{\ding{51}}\\

\cdashline{1-8}
GPT4RoI \cite{zhang2023gpt4roi} & \textcolor{igreen}{\ding{51}} & \textcolor{ired}{\ding{55}} & \textcolor{igreen}{\ding{51}} & \textcolor{igreen}{\ding{51}} & \textcolor{ired}{\ding{55}} & \textcolor{ired}{\ding{55}} & \textcolor{ired}{\ding{55}}\\
NExT-Chat \cite{zhang2023next} & \textcolor{igreen}{\ding{51}} & \textcolor{ired}{\ding{55}} & \textcolor{igreen}{\ding{51}} & \textcolor{igreen}{\ding{51}} & \textcolor{ired}{\ding{55}} & \textcolor{ired}{\ding{55}} & \textcolor{ired}{\ding{55}}\\
MiniGPT-v2 \cite{chen2023minigpt} & \textcolor{igreen}{\ding{51}} & \textcolor{ired}{\ding{55}} & \textcolor{igreen}{\ding{51}} & \textcolor{igreen}{\ding{51}} & \textcolor{ired}{\ding{55}} & \textcolor{ired}{\ding{55}} & \textcolor{ired}{\ding{55}}\\
Shikra \cite{chen2023shikra} & \textcolor{igreen}{\ding{51}} & \textcolor{ired}{\ding{55}} & \textcolor{igreen}{\ding{51}} & \textcolor{igreen}{\ding{51}} & \textcolor{ired}{\ding{55}} & \textcolor{ired}{\ding{55}} & \textcolor{ired}{\ding{55}}\\
Kosmos-2 \cite{peng2023kosmos} & \textcolor{igreen}{\ding{51}} & \textcolor{ired}{\ding{55}} & \textcolor{igreen}{\ding{51}} & \textcolor{igreen}{\ding{51}} & \textcolor{ired}{\ding{55}} & \textcolor{ired}{\ding{55}} & \textcolor{ired}{\ding{55}}\\
GLaMM \cite{rasheed2023glamm} & \textcolor{igreen}{\ding{51}} & \textcolor{ired}{\ding{55}} & \textcolor{igreen}{\ding{51}} & \textcolor{igreen}{\ding{51}} & \textcolor{ired}{\ding{55}} & \textcolor{ired}{\ding{55}} & \textcolor{ired}{\ding{55}}\\
Osprey \cite{yuan2023osprey} & \textcolor{igreen}{\ding{51}} & \textcolor{ired}{\ding{55}} & \textcolor{igreen}{\ding{51}} & \textcolor{igreen}{\ding{51}} & \textcolor{ired}{\ding{55}} & \textcolor{ired}{\ding{55}} & \textcolor{ired}{\ding{55}}\\
PixelLM \cite{ren2023pixellm} & \textcolor{igreen}{\ding{51}} & \textcolor{ired}{\ding{55}} & \textcolor{igreen}{\ding{51}} & \textcolor{igreen}{\ding{51}} & \textcolor{ired}{\ding{55}} & \textcolor{ired}{\ding{55}} & \textcolor{ired}{\ding{55}}\\
LLaVA-Plus \cite{liu2023llava} & \textcolor{igreen}{\ding{51}} & \textcolor{ired}{\ding{55}} & \textcolor{ired}{\ding{55}} & \textcolor{igreen}{\ding{51}} & \textcolor{igreen}{\ding{51}} & \textcolor{igreen}{\ding{51}} & \textcolor{ired}{\ding{55}}\\

\hline
VideoChat \cite{abs-2305-06355} & \textcolor{ired}{\ding{55}} & \textcolor{igreen}{\ding{51}} & \textcolor{ired}{\ding{55}} & \textcolor{ired}{\ding{55}} & \textcolor{ired}{\ding{55}} & \textcolor{ired}{\ding{55}} & \textcolor{ired}{\ding{55}}\\
Video-LLaMA \cite{abs-2306-02858}  & \textcolor{ired}{\ding{55}} & \textcolor{igreen}{\ding{51}} & \textcolor{ired}{\ding{55}} & \textcolor{ired}{\ding{55}} & \textcolor{ired}{\ding{55}} & \textcolor{ired}{\ding{55}} & \textcolor{ired}{\ding{55}}\\
Video-LLaVA \cite{lin2023video}  & \textcolor{igreen}{\ding{51}} & \textcolor{igreen}{\ding{51}} & \textcolor{ired}{\ding{55}} & \textcolor{ired}{\ding{55}} & \textcolor{ired}{\ding{55}} & \textcolor{ired}{\ding{55}}& \textcolor{ired}{\ding{55}} \\
Video-ChatGPT \cite{abs-2306-05424} & \textcolor{ired}{\ding{55}} & \textcolor{igreen}{\ding{51}} & \textcolor{ired}{\ding{55}} & \textcolor{ired}{\ding{55}} & \textcolor{ired}{\ding{55}} & \textcolor{ired}{\ding{55}} & \textcolor{ired}{\ding{55}}\\
GPT4Video \cite{wang2023gpt4video}  & \textcolor{ired}{\ding{55}} & \textcolor{igreen}{\ding{51}} & \textcolor{ired}{\ding{55}} & \textcolor{ired}{\ding{55}} & \textcolor{igreen}{\ding{51}} & \textcolor{ired}{\ding{55}} & \textcolor{ired}{\ding{55}}\\

\cdashline{1-8}
PG-Video-LLaVA \cite{munasinghe2023pg} & \textcolor{ired}{\ding{55}} & \textcolor{igreen}{\ding{51}} & \textcolor{igreen}{\ding{51}} & \textcolor{igreen}{\ding{51}} & \textcolor{ired}{\ding{55}} & \textcolor{ired}{\ding{55}}& \textcolor{ired}{\ding{55}} \\

\hline
NExT-GPT \cite{wu2023next} & \textcolor{igreen}{\ding{51}} & \textcolor{igreen}{\ding{51}} & \textcolor{ired}{\ding{55}} & \textcolor{ired}{\ding{55}} & \textcolor{igreen}{\ding{51}} & \textcolor{ired}{\ding{55}} & \textcolor{ired}{\ding{55}}\\

\cdashline{1-8}
\textsc{Vitron} (Ours) & \textcolor{igreen}{\ding{51}} & \textcolor{igreen}{\ding{51}} & \textcolor{igreen}{\ding{51}} & \textcolor{igreen}{\ding{51}} & \textcolor{igreen}{\ding{51}} & \textcolor{igreen}{\ding{51}} & \textcolor{igreen}{\ding{51}} \\

\bottomrule
\end{tabular}
\caption{
Comparisons of existing (partially, imperfect coverage) representative vision MLLM.
}
\vspace{-5mm}
\label{tab:comparisons-of-vision-LLM}
\end{table*}

\vspace{-2mm}
\section{Related Work}
\label{related_work}

\vspace{-3mm}

Achieving a profound understanding and comprehensive operational capabilities in vision, ranging from low-level visual pixel understanding \cite{CarionMSUKZ20,liu2023grounding,yang2022lavt,lai2023lisa,li2024omg,yang2022decoupling,wang2023look,li2022videoknet,li2023tube,li2023transformer} to high-level comprehension of overall semantics \cite{deng2009imagenet,krizhevsky2017imagenet,MilewskiMC20,fei2023scene,GuCWZLW23,ji2021improving,ji2022knowing,li2023variational,XiaoYL0JC22,fei2024enhancing,0004WXJC22,fei2024dysen}, represents a significant topic. 
Recent years have seen the development of highly potent large-scale vision models, such as ViT \cite{DosovitskiyB0WZ21} and CLIP \cite{RadfordKHRGASAM21}, which have achieved remarkable vision understanding capabilities; models like SAM \cite{kirillov2023segment} and SEEM \cite{zou2024segment} have solved vision segmentation tasks; and diffusion-based models \cite{abs-2102-05379,abs-2308-05095,mou2023t2i,abs-2212-05032,wu2023imagine,RombachBLEO22,fei2024video} have reached unprecedented performance in vision generation. 
Yet these models might lack an LLM as a central decision processor, unable to flexibly interpret user intent or execute tasks interactively \cite{abs-2305-11846,li2023fine,wu2023next}.
The emergence of LLMs has exhibited unprecedented intelligence capability \cite{chatgpt,abs-2210-11416,abs-2302-13971}. 
Extending the success of language understanding in LLMs, researchers have promptly investigated and developed various MLLMs, enabling LLMs to comprehend vision. 
By integrating high-performance vision encoders of images or videos into language-based LLMs, these models have been made capable of understanding vision signals \cite{gpt4,AlayracDLMBHLMM22,0008LSH23,qian2024momentor,abs-2304-08485}. 
Going beyond vision understanding, further research has aimed to enhance MLLMs, for instance, by endowing them with vision generation capabilities \cite{koh2023generating,sun2023generative} or supporting pixel-level understanding and grounding \cite{zhang2023gpt4roi,yuan2023osprey,ren2023pixellm,OMGLLaVA,wu2024towards2}. 
In Table \ref{tab:comparisons-of-vision-LLM} we summarize some existing popular vision MLLMs in terms of the vision function support.

However, we observe that current research on vision LLMs lacks depth in two critical aspects. 
Firstly, current vision LLMs tend to separate images and videos, supporting either one or the other. 
The construction of a unified MLLM is crucial, as vision inherently encompasses both static images and dynamic videos, both of which are core components of our visual world. 
Thus, covering both aspects simultaneously is essential for optimally adapting to practical applications. 
Although models like NExT-GPT \cite{wu2023next} have relatively well-supported unification across various modalities, they fall short in supporting pixel-level in-depth vision understanding and comprehensive support for vision operation tasks. 
The second issue is the incomplete support for vision tasks by existing MLLMs. 
Most current MLLMs primarily support understanding images or videos \cite{abs-2304-08485,abs-2304-10592}, with only a few supporting generation \cite{dong2023dreamllm,wang2023gpt4video} or editing/inpainting \cite{wu2024towards}. 
Building a generalist that can handle (almost) all vision-related tasks and operations in an end-to-end architecture should be the next major trend for vision MLLMs.
Yet simply integrating existing visual specialists into an LLM to form MLLMs is not sufficient enough, as genuine human-level AI should possess universal intelligence with robust cross-task generalizability \cite{morris2023levels}. 
Thus, it is necessary to further consider how to enable synergy effects \cite{dong2023dreamllm} among different task specialists within a generalist, for which goal, we have devised a synergy strategy in this work.
Besides, compared to the multimodal comprehension capabilities of MLLM, endowing MLLM with strong multimodal generative abilities is even more challenging. 
The key lies in how to effectively and unbiasedly convey MLLM's semantic understanding signals to the backbone decoder modules. 
There are two mainstream approaches to LLM-to-decoder message passing within the MLLM community. 
One is based on discrete textual instructions \cite{abs-2303-04671,abs-2303-17580,wang2024modaverse}, and the other on continuous signal embeddings \cite{koh2023generating,dong2023dreamllm,wu2023next}. 
However, we find that these two methods are complementary. 
Specifically, the former allows the LLM to efficiently convey task execution commands to the backend modules through simple text, but it struggles to provide modality-specific signals; the latter can conveniently carry the features needed for tasks, but fails to accurately convey execution intention (especially for managing many modules). 
In this work, we propose a hybrid method by integrating them together.

\vspace{-3mm}
\section{Architecture of \textsc{Vitron}}
\label{architecture}

\vspace{-3mm}

\textsc{Vitron} takes most common `\emph{encoder-LLM-decoder}' architecture paradigm, as in existing popular MLLMs \cite{abs-2304-08485,dong2023dreamllm,wu2023next}.
The overall framework is shown in Fig. \ref{fig:framework},
where three key blocks are included:
1) frontend vision\&language encoders, 2) central LLM for semantics understanding and text generation, and 3) backend decoder modules for user responding and vision manipulation.

\vspace{-3mm}
\subsection{Frontend Vision-Language Encoding}

\vspace{-2mm}
For both images and videos, we employ the CLIP ViT-L/14@336px \cite{RadfordKHRGASAM21} as the encoder, respectively. 
The video encoder independently processes each frame, further employing average pooling across the temporal dimension to yield overall temporal representation features.
Then, we employ a regional pixel-aware visual extractor as the sketch encoder for user interaction, e.g., clicking, drawing boxes or polygons, and making scribbles.
We mainly follow \cite{yuan2023osprey}, and use the object-based representations of mask regions that come from user's inputs, which not only encode the pixel-level visual features but also gather the spatial position information of each region. 
The region features are pooled with also the binary mask of spatial geometry of the object region encoded, and the resulting embeddings are used.
Then, the multimodal feature representations are passed to LLM via linear projection.

\vspace{-3mm}
\subsection{Core LLM}

\vspace{-2mm}

In \textsc{Vitron}, an LLM serves as the pivotal agent. 
Following the most common practice \cite{vicuna, abs-2305-16355, abs-2306-02858}, we utilize Vicuna (7B, version 1.5).
The LLM processes inputs from both language and visual modalities to perform semantic understanding and reasoning, and then make decisions. 
For visual comprehension tasks, LLM directly outputs textual responses for users.
On the other side, LLM also needs to transmit signals and instructions to backend modules, directing them to invocate more complex tasks that go beyond text generation, such as visual segmentation, generation, and editing. 
As emphasized earlier, the ability of LLMs to effectively and precisely convey messages is crucial to the performance of complex multimodal tasks. 
To this end, we propose fully integrating the advantages of the two common message-passing methods: \emph{discrete textual instructions} and \emph{continuous signal embeddings}. 
The former aids in accurately invoking different backbone modules (thanks to the LLM's proficiency in task dispatching), while the latter supplements with richer modality-preserved visual features that cannot be directly described through discrete text.
As depicted in Fig. \ref{fig:framework}, the LLM outputs 1) text responses for users, 2) text instructions for module invocation, and 3) feature embeddings of special tokens.
The feature embeddings are split into the task-specific features and the task-invariant fine-grained visual-language features.
Both the text instructions and feature embeddings are passed to backbone modules.

\vspace{-2mm}
\subsection{Backend Visual Specialists}
\vspace{-2mm}

To enable our MLLM with various visual task abilities, we integrate an array of singleton vision specialists into LLM.
For image generation and editing, we integrate the diffusion-based model GLIGEN \cite{li2023gligen}. 
For image and video segmentation, we opt for SEEM \cite{zou2024segment}. 
For video generation, ZeroScope \cite{zeroscope} and I2VGen-XL \cite{zhang2023i2vgen} are utilized for text-to-video and image-to-video tasks, respectively. 
Lastly, for video editing functionality, we incorporate StableVideo \cite{chai2023stablevideo}. 
The text instructions from LLM first determine which task module to invoke; simultaneously, feature embeddings are fed into the corresponding module's feature encoder to assist with task execution.
Specifically, we design a structured invocation template, including 1) Module name, 2) Invocation command, and 3) Region (optional) specifying a fine-grained vision feature needed for certain tasks.
The feature embeddings include both \emph{task-specific features} and \emph{task-invariant fine-grained features}. 
The purpose of this design is to achieve feature decoupling, during which we aim to have the task-invariant fine-grained features shared as widely as possible among all tasks to facilitate synergy between different tasks.

\vspace{-4mm}
\section{Pixel-aware Synergistic Vision-Language Understanding Tuning}
\label{sec:Pixel-aware Synergistic Vision-Language Understanding Tuning}

\vspace{-3mm}
With the \textsc{Vitron} framework, we now train the model with three stages of targets.
First, we try to endow it with basic multimodal capabilities, i.e., comprehension and generation.
Then, we engage in fine-grained vision grounding instruction tuning to further enhance the model's pixel-level perception abilities. 
Finally, we carry out cross-task synergy learning, maximizing the shared fine-grained features among all tasks.

\vspace{-3mm}
\subsection{Basic Multimodal Comprehension and Generation Skill Training}
\label{sec:Basic Multimodal Comprehension and Generation Skill Training}

\vspace{-2mm}
In the first stage of training, the primary goal is to equip the MLLM with basic multimodal understanding and generation abilities, including the frontend alignment of encoder-LLM, as well as the backend alignment of LLM-decoder.
Appendix $\S$\ref{Baisc MLLM Skill Training} details all the following three types of training.

\vspace{-3mm}
\paragraph{Overall Vision-Language Alignment Learning.}

This is to ensure the input vision and language are mapped to a unified feature space. 
Following prior common practice, we utilize datasets comprising `image-caption' pairs (CC3M \citep{SoricutDSG18}), `video-caption' pairs (Webvid \citep{BainNVZ21}), and `region-caption' pairs (RefCOCO \cite{kazemzadeh2014referitgame}) drawn from existing established corpora and benchmarks. 
When provided with an image, video, or specific visual region, we engage the frozen LLM to generate a text description or caption that aligns with the reference caption.

\vspace{-3mm}
\paragraph{Text Invocation Instruction Tuning.}
This step of training aims to equip the system with the precise capability to execute commands, allowing the LLM to generate appropriate and correct invocation text instructions. 
To accomplish this, we collect a total of 55,000+ instruction tuning samples.

\vspace{-3mm}
\paragraph{Embedding-oriented Decoder Alignment Tuning.}

Besides using explicit textual instruction to invocate downstream modules, the signal feature embedding/representation (from LLM) should also be fed to the modules.
Following \cite{wu2023next}, we align the feature embedding with all the visual modules' input encoders via the decoding-side projection layers, i.e., by minimizing their distances.

\vspace{-3mm}
\subsection{Fine-grained Spatiotemporal Vision Grounding Instruction Tuning}
\label{Fine-grained Spatiotemporal Vision Grounding Instruction Tuning}

\vspace{-3mm}
A visual generalist should require a strong capability of pixel-aware vision understanding of both images and videos.
Thus, we propose a fine-grained spatiotemporal vision grounding instruction tuning for \textsc{Vitron}. 
The core idea is to enable the LLM to ground the fine-grained spatiality of images and the detailed temporality of videos.
Appendix $\S$\ref{extesion:Fine-grained Spatiotemporal Vision Grounding Instruction Tuning} extends more detailed descriptions of the following three learning aspects.

\vspace{-3mm}
\paragraph{Image Spatial Grounding.}
Considering that the LLM alone can only output text, we design it to respond with the corresponding bounding box areas. 
We focus on two types of tasks: grounded image captioning \cite{ZhangSTXYZ21,ZhouWLHZ20} and referring image segmentation \cite{kazemzadeh2014referitgame}.

\vspace{-3mm}
\paragraph{Video Spatial-Temporal Grounding.}
For videos, the LLM must identify spatial regions and ground them within the temporal context of the video, essentially achieving video tracking. 
Similarly, we explore tasks such as grounded video captioning \cite{ZhouKCCR19} and referring video tracking \cite{WuHWD0S23}.

\begin{wrapfigure}{r}{0.47\textwidth}
\vspace{-6mm}
\begin{center}
    \includegraphics[width=0.47\textwidth]{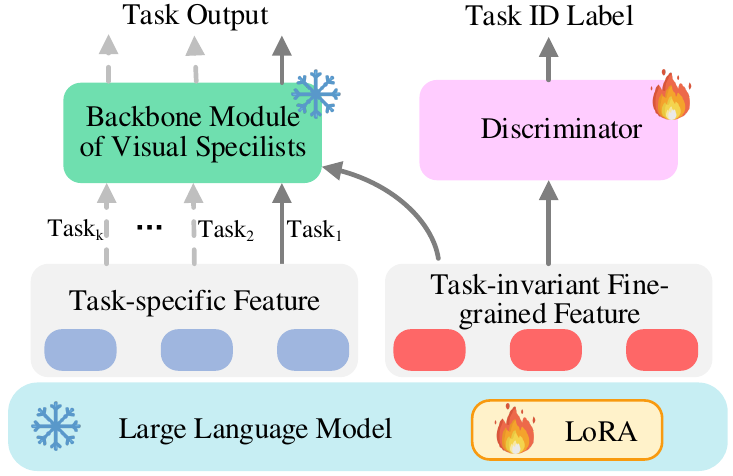}
\end{center}
\vspace{-4mm}
\caption{Illustration of the synergy module.}
\label{fig:synergy-module}
\vspace{-5mm}
\end{wrapfigure}

\vspace{-3mm}
\paragraph{Grounding-aware Vision QA.}
The grounding tasks mentioned above only touch upon the low-level aspects of vision perception. 
However, in many scenarios, it's essential for the LLM to possess high-level, in-depth vision reasoning capabilities, building upon the foundational low-level pixel grounding. 
Thus, we further introduce grounding-aware vision QA, including Image-QA \cite{schwenk2022okvqa,hudson2019gqa} and Video-QA \cite{yu2019activitynet}, enabling LLM to undertake semantic-level QA tasks based on the grounded results.

\vspace{-2mm}
\subsection{Cross-task Synergy Learning}
\label{Cross-task Synergy Learning}

\vspace{-2mm}

As a generalist, directly invoking different specialists leads to a critical issue: \emph{how to ensure that the different modules (tasks) work together synergistically?} 
Otherwise, without such collaboration, integrating them into a single compound system would be meaningless.
To achieve this, here we propose decomposing the signal feature embeddings into task-specific features and task-invariant fine-grained features. 
Intuitively, since all the visual tasks we focus on are fine-grained, the more extensively the task-invariant fine-grained features are shared among different tasks, the more these tasks can benefit from each other, thus gaining greater synergy.
Thereafter, we introduce a cross-task synergy learning module, as shown in Fig. \ref{fig:synergy-module}. 
We employ adversarial training \cite{BaiL0WW21} to decouple task-specific from task-invariant features. 
We first let different backbone visual specialists make task predictions based on these two features (via concatenation). 
Meanwhile, we encourage a third-party discriminator (acts as a classifier) to determine which is the current task based solely on the shared feature representation. 
Ideally, once the discriminator can no longer accurately identify the task, the shared feature can be considered the most purified and broadly applicable across tasks.

\vspace{-3mm}
\section{Experiments}
\label{experiments}

\vspace{-2mm}
Now we try to quantify the performance of \textsc{Vitron} on the four vision task groups, covering 12 tasks across 22 datasets.
All the training of \textsc{Vitron} is conducted on 10$\times$ A100 (80G) GPUs. 
To ensure a fair comparison, all subsequent experiments adopt settings same/similar to those of baseline systems, with evaluations following established practices. 
See more implementation details in Appendix $\S$\ref{Extended Details of Experimental Settings}.
Due to space limits, more experimental results are presented in Appendix $\S$\ref{More Experiment Results}.

\vspace{-2mm}
\subsection{Results on Vision Segmentation}

\begin{table*}[!h]
\vspace{-3mm}
\centering
\fontsize{8}{9}\selectfont
\setlength{\tabcolsep}{3.8mm}
\begin{tabular}{lcccccccc}
\hline
\multirow{2}{*}{\bf  Method }& \multicolumn{3}{c}{\bf RefCOCO \cite{kazemzadeh2014referitgame}}& \multicolumn{3}{c}{\bf RefCOCO+ \cite{yu2016modeling}} & \multicolumn{2}{c}{\bf RefCOCOg \cite{mao2016generation}} \\ 
\cmidrule(r){2-4} \cmidrule(r){5-7}  \cmidrule(r){8-9} 
& Val& 	TestA& 	TestB& 	Val& 	TestA& 	TestB& 	Val& 	Test \\
\hline
LAVT \cite{yang2022lavt}&	72.7&	75.8&	68.8&	62.1&	68.4&	55.1&	61.2&	62.1 \\
GRES \cite{liu2023gres}&	73.8&	76.5&	70.2&	66.0&	71.0&	57.7&	65.0&	66.0 \\
LISA \cite{lai2023lisa}&	74.1&	76.5&	71.1&	62.4&	67.4&	56.5&	66.4&	68.5 \\
NExT-Chat \cite{zhang2023next}&74.7& 78.9&	69.5&	65.1&	71.9&	56.7&	67.0&	67.0 \\

\rowcolor{lightgreen} \textsc{Vitron}&	\bf 75.5&	\bf 79.5&	\bf 72.2&	\bf 66.7&	\bf 72.5&	\bf 58.0&	\bf 67.9&	\bf 68.9 \\
\quad w/o syng.& -2.4 &	-2.0&	-1.9&	-1.7&	-2.1&	-1.5&	-1.8&	-1.6 \\

\hline
\end{tabular}
\caption{
Results (cIoU) of referring image segmentation.
`w/o syng.': without synergy learning.
}
\vspace{-3mm}
\label{tab:referring-expression-segmentation}
\end{table*}

\vspace{-2mm}

\paragraph{Image Segmentation.}
Table \ref{tab:referring-expression-segmentation} presents the results of referring image segmentation on three datasets: RefCOCO \cite{kazemzadeh2014referitgame}, RefCOCO+ \cite{yu2016modeling} and RefCOCOg \cite{mao2016generation}.
We compare with several significant models, including state-of-the-art non-MLLM approaches and the MLLM baseline, NExT-Chat. 
It is evident that our \textsc{Vitron}, while slightly underperforming compared to NExT-Chat on the RefCOCO Val\&TestA datasets, achieves superior performance on the remaining sets.

\begin{table}[!h]
\vspace{-3mm}
\begin{minipage}{\textwidth}
\centering
\begin{minipage}[b]{0.55\textwidth}
\fontsize{8}{9}\selectfont
\setlength{\tabcolsep}{1.2mm}
\centering
\setlength{\tabcolsep}{1mm}
\begin{tabular}{lcc}
\hline
\bf Method & \bf VidSTG \cite{zhang2020does}	 & \bf HC-STVG \cite{tang2021human} \\
\hline

G-DINO \cite{liu2023grounding} & 	25.3 & 	19.5 \\
Video-LLaMA \cite{abs-2306-02858} & 	28.6 & 	26.1 \\
Video-ChatGPT \cite{abs-2306-05424} & 	32.8 & 	20.8 \\
PG-Video-LLaVA \cite{munasinghe2023pg} & 	34.2 & 	28.3 \\

\rowcolor{lightgreen} \textsc{Vitron}	& \bf39.5 &	\bf31.4 \\
\quad w/o syng.& -4.3&	-3.7 \\
\hline
\end{tabular}
\vspace{1mm}
\captionof{table}{
\label{tab:video-spatial-grounding}
Results (mIoU) of video spatial grounding on two datasets.
}
\end{minipage}
\hfill
\begin{minipage}[b]{0.40\textwidth}
\centering
\fontsize{8}{9}\selectfont
\setlength{\tabcolsep}{1.8mm}
\begin{tabular}{lccc}
\hline
\multirow{1}{*}{\bf  Method }&  $\mathcal{J}$\&$\mathcal{F}$& 	$\mathcal{J}$& 	$\mathcal{F}$ \\

\hline
RDE \cite{li2022recurrent} &	77.4 &	73.6 &	81.2  \\
XMem \cite{cheng2022xmem} &	81.0  &	77.4 &	84.5  \\
DeAOT \cite{yang2022decoupling} & 	 80.7 &	76.9  &	 84.5  \\
ISVOS \cite{wang2023look} &	82.8 &	79.3 &  	86.2  \\

\rowcolor{lightgreen} \textsc{Vitron} &	\bf 84.2 &	\bf 81.5 &	\bf 86.7  \\
\quad w/o syng.& -2.1&	-1.3 & -1.0 \\

\hline
\end{tabular}
\vspace{1mm}
\captionof{table}{
\label{tab:video-object-Segmentation}
Results of video object segmentation on DAVIS 17 \cite{pont20172017} Test-Dev set.
}
\end{minipage}
\end{minipage}
\vspace{-5mm}
\end{table}

\vspace{-2mm}
\paragraph{Video Segmentation.}
For video segmentation, we explore two tasks: video spatial grounding (with bounding box) and video object segmentation (aka., video tracking; with mask). 
Table \ref{tab:video-spatial-grounding} showcases the comparisons between \textsc{Vitron} and current state-of-the-art (SoTA) video MLLMs in video spatial grounding. 
It is clear that \textsc{Vitron} significantly outperforms PG-Video-LLaVA. 
Table \ref{tab:video-object-Segmentation} presents a comparison of \textsc{Vitron} with some SoTA systems in video tracking, where our system continues to demonstrate superior performance.

\vspace{-2mm}
\subsection{Results on Fine-grained Vision Understanding}

\begin{wraptable}{r}{5.3cm}
\vspace{-13mm}
\fontsize{8}{9}\selectfont
\centering
\setlength{\tabcolsep}{1.7mm}
\begin{tabular}{lcc}
\hline
\bf Method & \bf METEOR & \bf CIREr \\
\hline
GRIT \cite{wu2022grit}	&	15.2 & 	71.6 \\   
Kosmos-2 \cite{peng2023kosmos}	&	14.1 &     	62.3 \\   
NExT-Chat \cite{zhang2023next}&	12.0 & 79.6 \\
MiniGPT-v2 \cite{chen2023minigpt}	&	15.0&	86.4 \\   
GLaMM \cite{rasheed2023glamm}	&	16.2&	106.0 \\   
Osprey \cite{yuan2023osprey}	&	16.6&	108.3 \\   
\rowcolor{lightgreen} \textsc{Vitron}	&	\bf 18.0 &	\bf111.6 \\  
\quad w/o syng.& -3.0&	-8.6 \\
\hline
\end{tabular}
\vspace{-2mm}
\caption{
\label{tab:regional-captioning}
Performance of image regional captioning on RefCOCOg \cite{mao2016generation}.
}
\vspace{-12mm}
\end{wraptable}

\vspace{-2mm}

Next, we evaluate \textsc{Vitron}'s capability in achieving fine-grained vision understanding, focusing mainly on region-level tasks for both images and videos.

\vspace{-2mm}
\paragraph{Region-level Image Understanding.}

We test \textsc{Vitron} on tasks including image referring expression comprehension and image regional captioning. 
The comparisons and results shown in Tables \ref{tab:regional-captioning} illustrate that \textsc{Vitron} surpasses the best baseline across various datasets and metrics, proving its strong and accurate fine-grained semantic understanding of images.

The above two tasks focus solely on the model's ability to recognize at the region level.
Taking a step further, we delve deeper into assessing the capability for image semantics understanding, particularly through image-based Visual Question Answering (VQA) tasks. 
These tasks effectively reflect the model's proficiency in comprehending the deeper semantic content of images. 
Table \ref{tab:image-VQA} displays the results across a series of six datasets for image-based VQA.
We primarily compare two groups of models: those with and without pixel-wise vision grounding capabilities. 
The findings indicate that models equipped with fine-grained grounding abilities indeed show stronger task performance, suggesting that fine-grained grounding contributes to a more profound understanding of semantics. 
Notably, our \textsc{Vitron} achieves the highest performance among the models evaluated.

\begin{table}[!h]
\vspace{-3mm}
\begin{minipage}{\textwidth}
\centering
\begin{minipage}[b]{0.48\textwidth}
\fontsize{8}{9}\selectfont
\setlength{\tabcolsep}{1.mm}
\centering
\begin{tabular}{lcccc}
\hline

\bf Method & \bf Ground?& 	\bf OKVQA \cite{schwenk2022okvqa} & \bf GQA \cite{hudson2019gqa} \\
\hline

Flamingo \cite{AlayracDLMBHLMM22} & 	\textcolor{ired}{\ding{55}}	 &    44.7 &   	  -  \\
BLIP-2 \cite{0008LSH23} & 	\textcolor{ired}{\ding{55}} & 	45.9 &  41.0  \\
InstructBLIP \cite{abs-2305-06500} &  	\textcolor{ired}{\ding{55}} & 	-  & 	49.5   \\
MiniGPT-4 \cite{abs-2304-10592} & 	\textcolor{ired}{\ding{55}} &  37.5  & 30.8	  \\
LLaVA \cite{abs-2304-08485} & 	\textcolor{ired}{\ding{55}} & 	54.4 &  41.3   \\
Shikra \cite{chen2023shikra} & 	\textcolor{igreen}{\ding{51}} & 	 47.2 &  -   \\
MiniGPT-v2 \cite{chen2023minigpt} &  	\textcolor{igreen}{\ding{51}} & 	 57.8  & 	 60.1  \\

\rowcolor{lightgreen} \textsc{Vitron} & 	\textcolor{igreen}{\ding{51}} & \bf 59.4 & \bf 	62.1  \\
\quad w/o syng.& \textcolor{igreen}{\ding{51}} & -2.0&	-1.7 \\

\hline
\end{tabular}
\vspace{1mm}
\captionof{table}{
\label{tab:image-VQA}
Results (accuracy) on image-based VQA.
}
\end{minipage}
\hfill
\begin{minipage}[b]{0.48\textwidth}
\centering
\fontsize{8}{9}\selectfont
\setlength{\tabcolsep}{.7mm}
\begin{tabular}{lccc}
\hline
\multirow{2}{*}{\bf  Method }&\multirow{2}{*}{\bf  Ground? }& \multicolumn{2}{c}{\bf ActivityNet-QA \cite{yu2019activitynet}} \\ 
\cmidrule(r){3-4} 
 & & \bf Accuracy& \bf Score \\
\hline
VideoChat \cite{abs-2305-06355} &	\textcolor{ired}{\ding{55}} &	  - &	2.2 \\
LLaMA-Adapter \cite{gao2023llama} &	\textcolor{ired}{\ding{55}} &	34.2 &	2.7 \\
Video-LLaMA \cite{abs-2306-02858}  &	\textcolor{ired}{\ding{55}}  &12.4	 &1.1 \\
Video-ChatGPT \cite{abs-2306-05424} &	\textcolor{ired}{\ding{55}} & 35.2 &	2.7 \\
Video-LLaVA \cite{lin2023video}  &	\textcolor{ired}{\ding{55}}	 &  45.3 &	3.3 \\
PG-Video-LLaVA \cite{munasinghe2023pg} &	\textcolor{igreen}{\ding{51}} &	39.9  &	3.3 \\

\rowcolor{lightgreen} \textsc{Vitron} &	\textcolor{igreen}{\ding{51}} &	\bf51.0 &	\bf	3.7 \\
\quad w/o syng.& \textcolor{igreen}{\ding{51}} & -4.4 &	-0.6 \\

\hline
\end{tabular}
\vspace{1mm}
\captionof{table}{
\label{tab:video-QA}
Results (accuracy and confidence Score) on video QA.
}
\end{minipage}
\end{minipage}
\vspace{-6mm}
\end{table}

\vspace{-2mm}
\paragraph{Region-level Video Understanding.}
Similarly, for videos, we evaluate the Region-level Video Understanding capability. Building on observations from images, we now directly engage in video QA tasks. 
Table \ref{tab:video-QA} presents the results on video QA across four representative datasets. 
Interestingly, while PG-Video-LLaVA has video grounding capabilities, it does not show better results than Video-LLaVA, which lacks grounding. However, our \textsc{Vitron} achieves superior performance. 
This indirectly proves that our system possesses more accurate video grounding capabilities (as previously demonstrated in Table \ref{fig:video-Segmentation}), aiding in better video semantics understanding.

\vspace{-2mm}
\subsection{Results on Vision Generation}

\begin{table}[!h]
\vspace{-3mm}
\begin{minipage}{\textwidth}
\centering
\begin{minipage}[b]{0.25\textwidth}
\fontsize{8}{8.5}\selectfont
\setlength{\tabcolsep}{1.mm}
\centering
\begin{tabular}{lc}
\hline
\bf Method & \bf FID ($\downarrow$) \\
\hline
GLIDE \cite{NicholDRSMMSC22} & 	12.24 \\
SD \cite{RombachBLEO22} & 11.21 \\
NExT-GPT \cite{wu2023next} & 11.28\\
Emu \cite{sun2023generative} & 11.66 \\
GILL \cite{koh2023generating} & 12.20 \\
DreamLLM \cite{dong2023dreamllm} & 8.46 \\
\rowcolor{lightgreen} \textsc{Vitron} &	\bf 7.57 \\
\quad w/o syng.&  +4.4 \\
\hline
\end{tabular}
\vspace{1mm}
\captionof{table}{
\label{tab:T2I-res}
Text-to-Image generation on COCO-caption data \cite{LinMBHPRDZ14}.
}
\end{minipage}
\hfill
\begin{minipage}[b]{0.35\textwidth}
\centering
\fontsize{8}{10}\selectfont
\setlength{\tabcolsep}{1.mm}
\begin{tabular}{lcc}
\hline
\bf Method & \bf FID ($\downarrow$) & \bf CLIPSIM ($\uparrow$) \\
\hline
CogVideo \cite{abs-2205-15868}&	23.59 & 	0.2631 \\   
MakeVideo \cite{abs-2209-14792} & 	13.17 & 	0.3049 \\   
Latent-VDM \cite{RombachBLEO22} &  14.25 & 	0.2756 \\   
Latent-Shift \cite{abs-2304-08477} & 15.23 & 	0.2773 \\
CoDi \cite{abs-2305-11846} & --- &  0.2890 \\
NExT-GPT \cite{wu2023next} & 13.04 &  0.3085 \\    

\rowcolor{lightgreen} \textsc{Vitron} &	\bf 10.11 &	\bf	0.3682 \\
\quad w/o syng.&  +3.17 &	-0.5672 \\

\hline
\end{tabular}
\vspace{1mm}
\captionof{table}{
\label{tab:T2V-res}
Text-to-Video generation on MSR-VTT \cite{XuMYR16}.
}
\end{minipage}
\hfill
\begin{minipage}[b]{0.32\textwidth}
\centering
\fontsize{8}{10}\selectfont
\setlength{\tabcolsep}{1.mm}
\begin{tabular}{lcc}
\hline
\bf Method & \bf FVD ($\downarrow$) & \bf IS ($\uparrow$) \\
\hline
AnimateAny \cite{dai2023fine}	&642.64&	63.87 \\   
DynamiCrafter \cite{xing2023dynamicrafter}&	404.50 &	41.97 \\   
SEINE \cite{chen2023seine}	&306.49&	 54.02 \\   
VideoCrafter1 \cite{chen2023videocrafter1}&	297.62&	50.88 \\  

\rowcolor{lightgreen} \textsc{Vitron} &	\bf 175.46 &	\bf 56.89 \\   
\quad w/o syng.&  +96.24&	-5.03 \\
\hline
\end{tabular}
\vspace{2mm}
\captionof{table}{
\label{tab:I2V-res}
Image-to-Video generation on UCF101 \cite{soomro2012ucf101}.
}
\end{minipage}
\end{minipage}
\vspace{-4mm}
\end{table}

\vspace{-2mm}
Next, we assess our system's capabilities in vision generation, focusing on three of the most representative types of generation tasks: text-to-image generation, text-to-video generation, and image-to-video generation. 
These tasks broadly cover the spectrum of image generation requirements. 
Tables \ref{tab:T2I-res}, \ref{tab:T2V-res}, and \ref{tab:I2V-res} showcase how our \textsc{Vitron} performs in comparison to other SoTA systems, including both MLLM and non-MLLM synthesizers.
The results clearly demonstrate that \textsc{Vitron} outperforms on all three tasks. 
For instance, in both text-to-image and text-to-video generation tasks, \textsc{Vitron} shows more advanced performance compared to NExT-GPT. 
Similarly, in the image-to-video generation task, \textsc{Vitron} still outshines the SoTA baseline, VideoCrafter1, showcasing superior results.

\vspace{-3mm}
\subsection{Results on Vision Editing}

\vspace{-2mm}
\paragraph{Image Editing.}

We use the MagicBrush dataset \cite{zhang2024magicbrush}, which challenges models with an editing query that demands a series of complex edits to an image. 
These edits include removing, changing, inpainting, and adding elements. 
Since there are currently no MLLM systems that support image editing, our comparison is limited to non-LLM expert systems. 
In Table \ref{tab:image-editing}, we present the performance of different models across various metrics. 
\textsc{Vitron} demonstrates stronger performance on all metrics, indicating its stable image editing capabilities.

\begin{table}[!h]
\vspace{-3mm}
\begin{minipage}{\textwidth}
\centering
\begin{minipage}[b]{0.6\textwidth}
\fontsize{8}{9}\selectfont
\setlength{\tabcolsep}{1mm}
\centering
\begin{tabular}{lccccc}
\hline
\bf Method & \bf CLIP$_{dir}$ ($\uparrow$) & \bf CLIP$_{img}$ ($\uparrow$)  & \bf CLIP$_{out}$ ($\uparrow$) & \bf L1 ($\downarrow$) \\
\hline

InstructPix2Pix \cite{brooks2023instructpix2pix}  & 	 0.115  &  	 0.837  &  	 0.245  &   0.093  \\
MagicBrush \cite{zhang2024magicbrush}  & 	  0.123   &  	0.883  &  	 0.261  &  	 0.058 \\
PnP \cite{tumanyan2023plug}  & 	 0.025   & 	  0.568   & 	 0.101  &  	  0.280   \\
NT-Inv \cite{mokady2023null}  & 	 0.121   & 	 0.752   & 	 0.263   & 	 0.077   \\
Emu-Edit \cite{sheynin2023emu}  &  	0.135  &  	 0.897  & 	 0.261  &  0.052 \\

\rowcolor{lightgreen} \textsc{Vitron}  &  \bf 	0.142  &  \bf 	0.910  & 	 \bf 0.274  &  \bf  0.047  \\
\quad w/o syng.&  -0.012 &	-0.104 & -0.078 & + 0.036\\

\hline
\end{tabular}
\vspace{1mm}
\captionof{table}{
\label{tab:image-editing}
Image editing results on MagicBrush \cite{zhang2024magicbrush}.
} 
\end{minipage}
\hfill
\begin{minipage}[b]{0.37\textwidth}
\centering
\fontsize{8}{11}\selectfont
\setlength{\tabcolsep}{0.7mm}
\begin{tabular}{lp{1cm}p{1.3cm}}
\hline
\bf Method & \bf Target-Editing & \bf NonTarget-Unediting \\
\hline

Text2LIVE \cite{bar2022text2live}	&\multicolumn{1}{c}{4.5}	&\multicolumn{1}{c}{1.3}  \\   
Tune-A-Video \cite{abs-2212-11565}	&\multicolumn{1}{c}{7.8}&	\multicolumn{1}{c}{4.6}  \\   

\rowcolor{lightgreen} \textsc{Vitron}	&\multicolumn{1}{c}{\bf 8.9}	&\multicolumn{1}{c}{\bf 8.2}  \\   
\quad w/o syng.&  \multicolumn{1}{c}{-2.2} &	\multicolumn{1}{c}{-1.6} \\

\hline
\end{tabular}
\vspace{1mm}
\captionof{table}{
\label{tab:video-editing}
Human evaluation on video editing.
}
\end{minipage}
\end{minipage}
\vspace{-5mm}
\end{table}

\vspace{-4mm}
\paragraph{Video Editing.}

For video editing, the community currently lacks a standardized benchmark and evaluation method akin to those for image editing. Therefore, we opted for a manual evaluation approach. We asked different video editing systems to edit the same video based on the same query, after which five individuals were asked to score the edited videos. The evaluation focused on 1) the success of target content modifications and 2) the faithfulness/fidelity of non-target content.
Table \ref{tab:video-editing} presents the manual evaluation results for video editing. 
It is clear that \textsc{Vitron} outperforms the two baseline systems in both respects, showcasing superior video editing capabilities. 
Following this, we visualized the process of video editing by \textsc{Vitron}.

\vspace{-3mm}
\section{Discussions}

\vspace{-2mm}
Above we demonstrate the overall efficacy of \textsc{Vitron} via extensive quantitative comparison.
Now we take one step further, exploring how and why the system advances via in-depth analyses.

\vspace{-2mm}
\paragraph{$\blacktriangleright$ Discrete Textual Instruction or Continuous Signal Embedding, Which Better?}

\begin{figure}[!h]
  \vspace{-1mm}
\centering
\includegraphics[width=1\columnwidth]{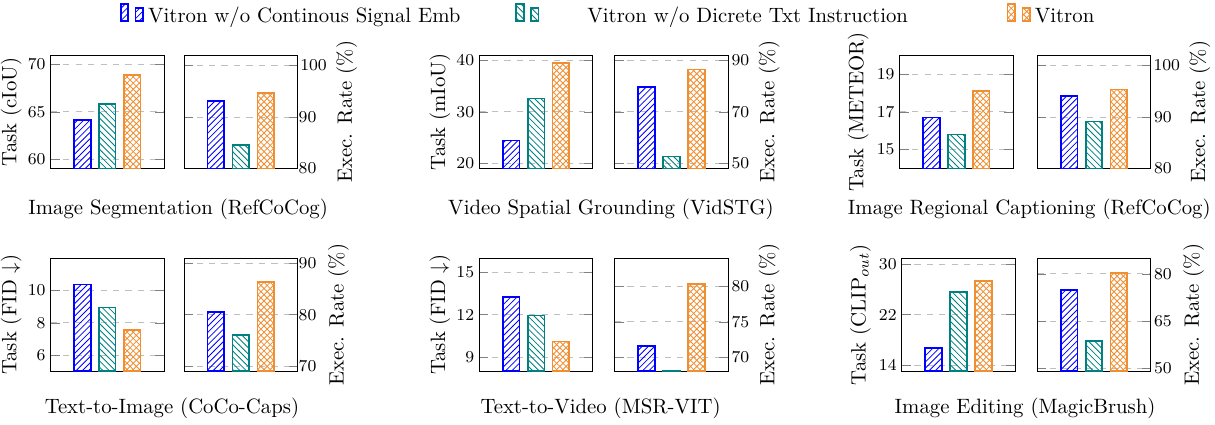}
\vspace{-4mm}
\caption{
The influences of using different strategies for message passing.
}
  \vspace{-2mm}
\label{fig:msgpass}
\end{figure}

Firstly, we explore different message-passing mechanisms to determine whether discrete textual instruction is more beneficial, or whether continuous signal embedding is better for building a multi-modal generalist. 
Also, we validate the pros and cons of the proposed hybrid method of message passing. 
We conduct tests on 6 tasks, where we compare the task performance of \textsc{Vitron} using the hybrid method (default setting), without signal embedding and without text instruction, as well as the successful execution rate of the backend task module. 
Fig. \ref{fig:msgpass} presents the results. 
As can be observed, overall, the performance under scenarios utilizing both methods is consistently better, which confirms the effectiveness of our hybrid mode. 
Meanwhile, we find that the method of text instruction is more conducive to the successful execution of backend modules, but soft feature embedding seems to be more useful in terms of specific task performances.

\vspace{-2mm}
\paragraph{$\blacktriangleright$ How Much Does Each Fine-grained Visual Grounding Learning Contribute?}

\begin{figure}[!t]
  \vspace{-1mm}
\centering
\includegraphics[width=1\columnwidth]{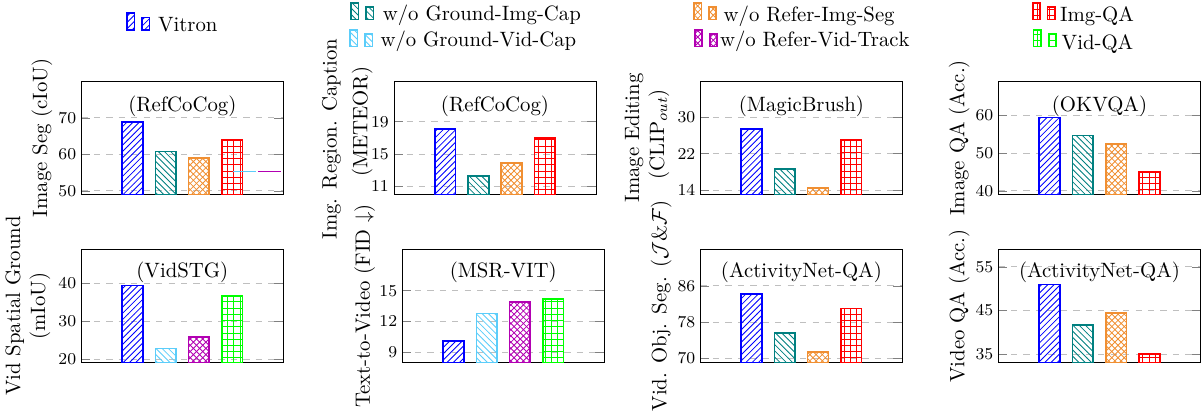}
\vspace{-4mm}
\caption{
The impact of various fine-grained visual grounding learning strategies.
}
\label{fig:fg-vl}
\end{figure}

Next, we validate the specific contribution of the various fine-grained visual grounding learning strategies proposed in $\S$\ref{Fine-grained Spatiotemporal Vision Grounding Instruction Tuning}. 
Fig. \ref{fig:fg-vl} (the top 4 relate to image tasks, and the bottom 4 to video tasks) shows the impact on performance when a particular learning strategy is removed.
Generally, all these 3 types of fine-grained visual grounding learning strategies are vital for different downstream tasks. 
For instance, grounding and referring segmentation tasks directly influence fine-grained visual recognition tasks, whereas tuning for grounding-aware visual QA considerably boosts cognition level QA tasks. 
This verifies the efficacy of our proposed fine-grained visual grounding tuning strategies.

\begin{figure}[!t]
\centering
\includegraphics[width=1\columnwidth]{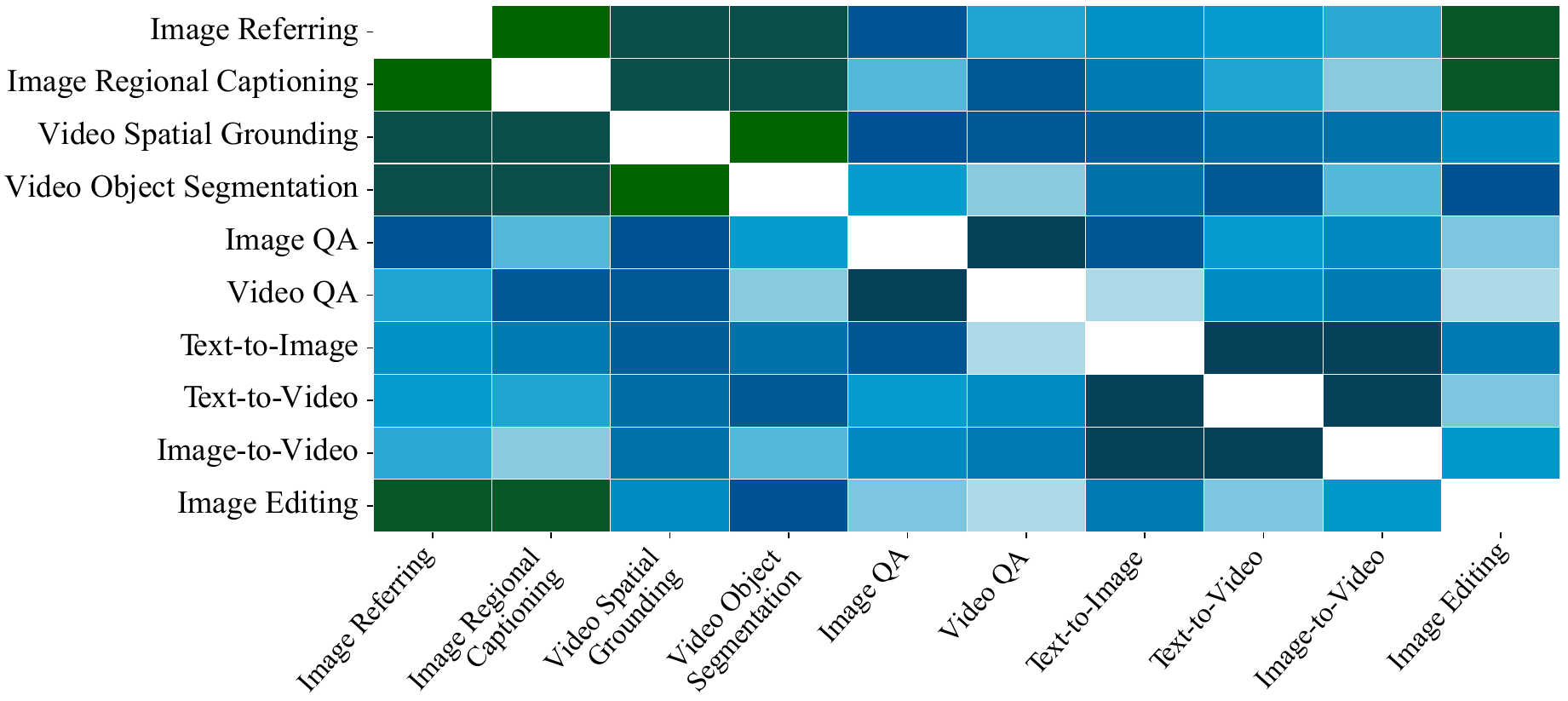}
\vspace{-4mm}
\caption{
The synergy correlation between each pair of visual tasks.
The deeper the color of the cell, the more synergistic they are in between.
}
  \vspace{-2mm}
\label{fig:synergy-heatmap}
\end{figure}

\vspace{-2mm}
\paragraph{$\blacktriangleright$Does \textsc{Vitron} Really Achieve Cross-task Synergy?}
Finally, we investigate if our system could adequately support cross-task synergy. 
Based on the results of the ablation item for the `synergy module' in Table \ref{tab:referring-expression-segmentation} to Table \ref{tab:video-editing}, we can observe that the synergy learning mechanism indeed positively influences overall performance. 
In Fig. \ref{fig:synergy-heatmap} we further study whether there is synergy between different tasks and their collaborative relations.
For ease of study, we considered a one-to-one mapping relationship, studying the cooperation between pairs of tasks one at a time. 
It is evident that the cooperative effects vary between different tasks.
Tasks or backbone modules that rely more heavily on fine-grained visual features gained more significant improvements. 
This also demonstrates that our synergy learning module can successfully facilitate cross-task synergy.

\vspace{-2mm}
\section{Conclusion}

\vspace{-2mm}
In this work, we present \textsc{Vitron}, a grand unified pixel-level vision LLM for seamlessly understanding (perceiving and reasoning), generating, segmenting (grounding and tracking), and editing (inpainting) both images and videos.
We further introduce a novel hybrid method of message passing that combines discrete textual instructions with continuous signal embeddings to ensure precise function invocation. 
Furthermore, \textsc{Vitron} employs pixel-level spatiotemporal vision-language alignment to enhance its fine-grained visual capabilities. 
A cross-task synergy module is also developed to optimize the use of task-invariant fine-grained visual features, boosting synergy across various visual tasks. 
On 12 visual tasks across 22 datasets, \textsc{Vitron} exhibits extensive capabilities in visual segmentation, fine-grained vision understanding, generation, and editing.
Overall, this research showcases the great potential to build a vision-language generalist that can advance toward a more unified AI.

\vspace{-2mm}
\section*{Acknowledgements}

\vspace{-2mm}
This research is supported by Skywork AI, NExT++ Research Center,
and CCF-Kuaishou Large Model Explorer Fund, Project of Future High-tech Video Intelligent Technology Innovation Center.




\newpage

{\small
\bibliographystyle{plainnat}
\bibliography{ref}
}

\newpage

\appendix

\section{Details of Backbone Visual Modules/Specialists}
\label{details_of_visual_backbone}

To address the inability of text-based LLMs in handling various vision tasks, we consider integrating off-the-shelf external modules. 
Once the LLM generates invocation details through understanding the input and recognizing the user's intent, the corresponding modules are activated to produce non-textual outputs. 
Technically, we employ a variety of current SoTA expert models for vision processing.
For image generation and editing, we integrate the diffusion-based model GLIGEN \cite{li2023gligen}. 
For image and video segmentation, we opt for SEEM \cite{zou2024segment}. 
For video generation, ZeroScope \cite{zeroscope} and I2VGen-XL \cite{zhang2023i2vgen} are utilized for text-to-video and image-to-video tasks, respectively. 
Lastly, for video editing functionality, we incorporate StableVideo \cite{chai2023stablevideo}. 
In Table \ref{tab:backend-modules}, we summarize the functionality of each backend module, along with a specification of the inputs and outputs.

\begin{table*}[!h]
\centering
\fontsize{8}{10}\selectfont
\setlength{\tabcolsep}{3mm}
\begin{tabular}{cllll}
\hline

\bf No.& 	\bf Function & 	\bf Model& 	\bf Input& 	\bf Output \\
\hline

\rowcolor{imageblue} 1&	Text Generation&	-&	-	&- \\

\rowcolor{textred} 2&	Image Generation&	GLIGEN \cite{li2023gligen}&	Text&	Image \\
\rowcolor{textred} 3&	Image Segmentation&	SEEM \cite{zou2024segment}&	Text, Image&	Image, Mask | BBox \\
\rowcolor{textred} 4&	Image Editing&	GLIGEN \cite{li2023gligen}&	Text, Image [BBox | Mask]&	Image \\

\rowcolor{videogreen} 5&	&	ZeroScope \cite{zeroscope}&	Text&	Video \\
\rowcolor{videogreen} 6&	\multirow{-2}{*}{Video Generation}& I2VGen-XL \cite{zhang2023i2vgen}&	Image&	Video \\
\rowcolor{videogreen} 7&	Video Segmentation&	SEEM \cite{zou2024segment}&	Text, Video [BBox | Mask]&	Video, Mask | BBox \\
\rowcolor{videogreen} 8&	Video Editing&	StableVideo \cite{chai2023stablevideo}&	Text, Video&	Video \\

\hline
\end{tabular}
\caption{
Summary of backend modules in \textsc{Vitron}.
}
\vspace{-2mm}
\label{tab:backend-modules}
\end{table*}

\section{Extensions of Pixel-aware Synergistic 
Vision-Language Understanding Learning}

This section extends more details of the $\S$\ref{sec:Pixel-aware Synergistic Vision-Language Understanding Tuning} in the main article.

\vspace{-2mm}
\subsection{Baisc MLLM Skill Training}
\label{Baisc MLLM Skill Training}

\vspace{-2mm}
\paragraph{Overall Vision-Language Alignment Learning.}

In line with the methodologies in current MLLMs, our approach involves
This step aims at mapping the input vision language features to a unified feature space. 
This space creates representations that the central LLM can interpret, thereby enabling it to process incoming vision signals effectively. 
We utilize datasets of `image-caption' pairs (CC3M \citep{SoricutDSG18}), `video-caption' pairs (Webvid \citep{BainNVZ21}), and `region-caption' pairs (RefCOCO \cite{kazemzadeh2014referitgame}) from existing corpora. 
When provided with an image, video, or specific visual region, we engage the frozen LLM to generate a text description or caption that aligns with the reference caption.

\paragraph{Invocation-oriented Instruction Tuning.}

The above phase of training endows both the LLM and the frontend encoders with the ability to understand vision.
This step, instruction tuning for invocation, aims to equip the system with the precise capability to execute commands, allowing the LLM to generate appropriate and correct invocation text. 
This text is then used to trigger various backend task execution modules. 
Different terminal vision tasks might require distinct invocation commands.
To unify this, we try to standardize the LLM's response output into a structured text format, which includes:
1) User response output, which directly replies to the user's input.
2) Module name, indicating which function or task is to be executed.
3) Invocation command, a meta-instruction for triggering the task module.
4) Region (optional), specifying a fine-grained vision feature needed for certain tasks, such as in video tracking or vision editing, where backend modules require this information. For the region, based on LLM's pixel-level understanding, a bounding box described by the coordinates (X$_l$, Y$_t$, X$_r$, Y$_b$) will be output.
Following, we exemplify a video tracking example for the structured LLM response for module invocation.

\begin{tcolorbox}[breakable,fontupper=\linespread{0.9}\selectfont]
{
\small
$\blacktriangleright$ \textbf{Structured LLM response for module invocation (with video tracking example):}\\
\textbf{Input:} \videologo  \sketchlogo \\
Can you help me track the clock I circled in the video?\\

\textbf{A: [LLM response]}\\
Sure! You provide a video where there is a female cop talking on the walkie-talkie, responding to emergency calls, and crime prevention. Also in the scene, the clock is green in color. Following I will outline the clock in the video. \\

\textbf{B: [Task Instruction]}\\
\textbf{<Module>} Video Segmentation \textbf{</Module>}\\
\textbf{<Instruction>} segmentation: clock \textbf{</Instruction>}\\
\textbf{<Region>} (0.23, 0.35, 0.11, 0.26) \textbf{</Region>}\\
}
\end{tcolorbox}

\begin{table*}[!h]
\centering
\fontsize{8}{14}\selectfont
\setlength{\tabcolsep}{2.8mm}
\begin{tabular}{llllc}
\hline

\bf Function & 	\bf \textsc{Vitron} Input &  \bf Module Input & 	\bf Data Source	& 	\bf Amount \\
\hline

Image Generation&	\textlogo &	 Image-Caption &	CC3M \cite{SoricutDSG18} & 4,000 \\
\hline

\multirow{3}{*}{Image Segmentation} 
&	\textlogo, \sketchlogo, \imagelogo &  \imagelogo, \sketchlogo &	RefCOCO \cite{kazemzadeh2014referitgame} & 4,000 \\
\cdashline{2-5}
&	\textlogo, \sketchlogo, \imagelogo, \imagereflogo  &  \imagelogo, \imagereflogo, \sketchlogo &	RefCOCO \cite{kazemzadeh2014referitgame} & 5,000 \\
\cdashline{2-5}
&	\textlogo, \imagelogo &  \imagelogo, Object-Name &	gRefCOCO \cite{liu2023gres} & 2,028 \\
\hline

\multirow{2}{*}{Image Editing} 
&	\textlogo, \sketchlogo, \imagelogo & \imagelogo, \sketchlogo  &	COCO2017 \cite{LinMBHPRDZ14} &	4,000 \\
\cdashline{2-5}
&	\textlogo, \imagelogo & \imagelogo, Bounding-Box & MagicBrush \cite{zhang2024magicbrush} &	5,000 \\
\hline
\hline

\multirow{2}{*}{Video Generation}
& \textlogo &  Video-Caption &	WebVid \citep{BainNVZ21} &	7,000 \\
\cdashline{2-5}
& \textlogo, \imagelogo &  \imagelogo &	LAION-400M \cite{schuhmann2021laion} &	4,000 \\
\hline

\multirow{2}{*}{Video Segmentation}
& \textlogo, \sketchlogo, \videologo & \videologo, \imagereflogo, \sketchlogo &	WebVid \citep{BainNVZ21}, VG \cite{KrishnaZGJHKCKL17} &	5,000 \\
\cdashline{2-5}
& \textlogo, \videologo & \videologo, \imagereflogo, Bounding-Box &	WebVid \citep{BainNVZ21} &	5,000 \\
\hline

\multirow{2}{*}{Video Editing}
& \textlogo, \sketchlogo, \videologo & \videologo, Editing-Query &	WebVid \citep{BainNVZ21} &	5,000 \\
\cdashline{2-5}
& \textlogo, \videologo & \videologo, Editing-Query &	WebVid \citep{BainNVZ21} &	5,000 \\

\hline
\end{tabular}
\caption{
Feature summary of the constructed data for text invocation instruction tuning.
{\imagereflogosmall} in image segmentation means the reference image provided by users.
{\imagereflogosmall} in video segmentation means the intermediate referred video keyframe interpreted within the system.
}
\vspace{-2mm}
\label{tab:it-data}
\end{table*}

To teach the LLM to produce the correct invocation responses, we need to create data for instruction tuning. 
A key is ensuring that the data covers all possible scenarios.
We must take into account different ways users might interact with the system for each functionality mentioned in Table \ref{tab:backend-modules} (except for text generation). 
For example, when requesting video creation, a user might describe what they want purely in text, or provide a reference image as the basis for the desired video. 
Similarly, for editing images or videos, users could express their editing requests either through text, or by using sketches, scribbles and other interactions.
Thus, the LLM needs to be versatile in accepting various types of user inputs and generating an accurate invocation response that matches the requirements of the backend modules. 
Technically, we make use of the existing annotated datasets for various vision tasks included in this work.
For each task under specific different user input scenarios, with the corresponding data, we design various template dialogue-format examples.
Based on these examples we then prompt the GPT-4 to generate more samples under various topics and enriched scenarios.
Finally, we collect a total of 55,000+ invocation-oriented instruction tuning samples.
In Table \ref{tab:it-data} we provide a summary of these datasets.

\paragraph{Embedding-oriented Decoder Alignment Tuning.}

Besides using the explicit textual instruction to invocate downstream modules, also the signal feature embedding/representation (from LLM) should also be fed to the modules.
Denote the \emph{task-specific features} as $\bm{v}^p$
and \emph{task-invariant fine-grained features} as $\bm{v}^s$. 
We will concatenate them as one unified feature embedding $\bm{v} = [\bm{v}^p;\bm{v}^s]$ and then send $\bm{v}$ to the downstream module.

Following \cite{wu2023next}, we align the feature embedding with all the visual module's input encoders via the decoding-side projection layers.
We do this feature alignment learning by minimizing the distance between the projected feature embedding and the module's input encoder.
For example for diffusion-based image or video generation, we may directly use the textual condition encoder, while keeping all the other modules fixed.
Technically, to endow the model to produce other modalities beyond text, we add the signal tokens to the vocabulary of the LLM. 
In the alignment training phase, we mainly take the captions from CC3M, WebVid, and AudioCaps as inputs and concatenate them with the special signal tokens as outputs.
The loss function comprises three key components: 1) the negative log-likelihood of producing signal tokens, and 2) the caption alignment loss: $\mathit{l}_2$-distance between the hidden states of signal tokens produced by the LLM and the conditional text representations derived from the text encoder within diffusion models, and 3) conditional latent denoising loss \cite{RombachBLEO22}.

\subsection{Fine-grained Spatiotemporal Vision Grounding Instruction Tuning}
\label{extesion:Fine-grained Spatiotemporal Vision Grounding Instruction Tuning}

\vspace{-1mm}
We propose a fine-grained spatiotemporal vision grounding instruction tuning for \textsc{Vitron}. 
The core idea is to enable the LLM to ground the fine-grained spatiality of images and the detailed temporality of videos.
Technically, we leverage LoRA \cite{HuSWALWWC22} to enable a small subset of parameters within the LLM to be updated during the tuning.

\paragraph{Image Spatial Grounding.}

Considering that the LLM alone can only output text, we design it to respond with the corresponding bounding box areas. 
We focus on two types of tasks: grounded image captioning and referring image segmentation. 
For grounded image captioning, we input an image and identify a specific object within it, prompting the LLM to describe the identified object. 
Conversely, for referring image segmentation (where we consider outputting a bounding box, akin to phrase grounding), the task involves inputting a complete image along with a related phrase or sentence description, and the LLM is expected to output the object's spatial bounding box, represented by coordinate numbers (X$_l$, Y$_t$, X$_r$, Y$_b$).
The X and Y coordinates are normalized real values within the range [0, 1], where <X$_l$> and <Y$_t$> indicate the top-left corner, and <X$_r$> and <Y$_b$> represent the bottom-right corner of the bounding box.
We consider datasets such as Flickr30K Entities \cite{plummer2015flickr30k}, RefCOCO \cite{kazemzadeh2014referitgame}, RefCOCO+ \cite{yu2016modeling} and RefCOCOg \cite{mao2016generation}. 
Utilizing GPT-4, we preprocess and expand the original data into a multi-turn QA format, thereby generating the most suitable instruction-tuning data.


\begin{minipage}{0.58\textwidth} 
\begin{tcolorbox}[breakable,fontupper=\linespread{0.9}\selectfont,]
{
\small
$\blacktriangleright$ \textbf{Grounded Image Captioning:}\\
\textbf{Input:} \imagelogo \\
Please generate a detailed caption for the given image, and clearly link each part of the caption to specific objects or areas in the image which you can denote with a bounding box with ``object: (X$_l$, Y$_t$, X$_r$, Y$_b$)'' format.\\
\textbf{Output:}\\
\emph{A girl wearing a pink dress is sitting on a long bench in the park reading a book.}\\
\emph{girl: (X$_l$, Y$_t$, X$_r$, Y$_b$)},\\
\emph{pink dress: (X$_l$, Y$_t$, X$_r$, Y$_b$)}, ...
}
\end{tcolorbox}
\end{minipage}%
\begin{minipage}{0.42\textwidth} 
\begin{tcolorbox}[breakable,fontupper=\linespread{0.9}\selectfont,]
{
\small
$\blacktriangleright$ \textbf{Referring Image Segmentation:}\\
\textbf{Input:} \imagelogo\\
Please identify the target object from the given images based on the following text query:
``A man in a black shirt holding a water bottle''.\\
Please output the bounding box (X$_l$, Y$_t$, X$_r$, Y$_b$) of the target object.\\
\textbf{Output:}\\
\emph{(X$_l$, Y$_t$, X$_r$, Y$_b$)}
}
\end{tcolorbox}
\end{minipage}

\paragraph{Video Spatial-Temporal Grounding.}

For videos, the LLM must not only identify spatial regions but also ground these within the temporal context of the video, essentially achieving video tracking. 
Similarly, we explore tasks such as grounded video captioning and referring video tracking. While the approach is akin to that used for images, it requires not just the output of object bounding box coordinates in each frame but also the specification of the duration of the frames (F$_s$, F$_e$) in which these coordinates are to be outputted.
Here <F$_s$> denotes the starting frame number, and <F$_e$> means the ending frame number.
We mainly construct the required instruction tuning data from several key video tracking datasets: LaSOT \cite{fan2019lasot}, GOT10K \cite{huang2019got}, DanceTrack \cite{sun2022dancetrack}, and MOT16 \cite{milan2016mot16}.

\begin{minipage}{0.54\textwidth} 
\begin{tcolorbox}[fontupper=\linespread{0.9}\selectfont,]
{
\small
$\blacktriangleright$ \textbf{Grounded Video Captioning:}\\
\textbf{Input:} \videologo \\
Please generate a caption for the given video, and link each part of the caption to specific objects in the video with its temporal presence duration.
You should denote it with a bounding box with the starting and ending frame number in format as ``object: (X$_l$, Y$_t$, X$_r$, Y$_b$ | F$_s$, F$_e$)''.\\
\textbf{Output:}\\
\emph{A dog is running after a ball on the grass.}\\
\emph{dog: (X$_l$, Y$_t$, X$_r$, Y$_b$ | F$_s$, F$_e$)},\\
\emph{ball: (X$_l$, Y$_t$, X$_r$, Y$_b$ | F$_s$, F$_e$)}, ...
}
\end{tcolorbox}
\end{minipage}%
\begin{minipage}{0.46\textwidth} 
\begin{tcolorbox}[fontupper=\linespread{0.9}\selectfont,]
{
\small
$\blacktriangleright$ \textbf{Referring Video Tracking:}\\
\textbf{Input:} \videologo\\
Please track the specified object throughout the video based on the following given description, and mark its starting and ending position in each frame:
``A man is skiing on a thick-snow-covered slope.''\\
Please output in the format of ``(X$_l$, Y$_t$, X$_r$, Y$_b$ | F$_s$, F$_e$)''.\\
\textbf{Output:}\\
\emph{(X$_l$, Y$_t$, X$_r$, Y$_b$ | F$_s$, F$_e$)}
}
\end{tcolorbox}
\end{minipage}


\paragraph{Grounding-aware Vision QA.}

The grounding tasks mentioned above only touch upon the low-level aspects of vision perception. 
However, in many scenarios, it's essential for the LLM to possess high-level, in-depth vision reasoning capabilities, building upon the foundational low-level pixel grounding. 
Therefore, we further introduce grounding-aware vision QA, enabling the LLM to undertake semantic-level QA tasks based on the grounded results. 
Specifically, we consider both image-grounded QA and video-grounded QA. 
The data sources for these tasks remain the same as mentioned above for the grounding tasks. 
And we again rely on GPT-4\footnote{\url{https://openai.com/index/gpt-4/}} to use its imaginative capabilities to design reasonable questions and answers based on the given grounding content of the above grounding/segmentation datasets, formatting these into an instruction-tuning format.

\begin{minipage}{0.42\textwidth} 
\begin{tcolorbox}[fontupper=\linespread{0.9}\selectfont,]
{
\small
$\blacktriangleright$ \textbf{Grounded Image QA:}\\
\textbf{Input:} \imagelogo \\
Based on the given image, please select the correct answer among all the candidates:
Q: Where is the child sitting? A: 1) arms, 2) ground, 3) chair, 4) bed.\\
Please first identify and ground the target object (in coordinates) mentioned in the question, and then proceed to answer the question.\\
\textbf{Output:}\\
The target object mentioned in the question is "child," with the position given by \emph{(X$_l$, Y$_t$, X$_r$, Y$_b$)}. From this, it can be determined that the child is being held by a woman, which means the child is sitting in the woman's arms. Therefore, the answer is 1) arms.
}
\end{tcolorbox}
\end{minipage}%
\begin{minipage}{0.58\textwidth} 
\begin{tcolorbox}[fontupper=\linespread{0.9}\selectfont,]
{
\small
$\blacktriangleright$ \textbf{Grounded Video QA:}\\
\textbf{Input:} \videologo \\
Based on the provided video, answer the following question by choosing the most appropriate answer from the options given.
Q: If the rope in the video breaks, what would happen to the person in the video? A: 1) running, 2) jumping up, 3) flying away, 4) fracture.\\
First, analyze the spatial position and temporality of the target object mentioned in the question within the video, and based on this analysis, determine the answer to the question.\\
\textbf{Output:}\\
The objects involved in the question are the rope \emph{(X$_l$, Y$_t$, X$_r$, Y$_b$ | F$_s$, F$_e$)} and the person \emph{(X$_l$, Y$_t$, X$_r$, Y$_b$ | F$_s$, F$_e$)}, with their relationship being that the person is climbing with the assistance of the rope.
Based on common sense, if the rope were to break, the person would fall to the ground from a height. The most likely outcome is that the person would break a bone.
}
\end{tcolorbox}
\end{minipage}


\subsection{Cross-task Synergy Learning}

Let's denote the \emph{task-specific features} as $\bm{v}^p$
and \emph{task-invariant fine-grained features} as $\bm{v}^s$. 
We will concatenate them as one unified feature embedding $\bm{v} = [\bm{v}^p;\bm{v}^s]$ and then send $\bm{v}$ to the downstream module $M_k$ for the $k$-th task prediction:
\begin{equation}
    y_k \leftarrow \text{M} (\bm{v}) .
\end{equation}
Although we separate the two feature spaces into the shared one $\bm{v}^s$ and private one $\bm{v}^p$, there are still chances that the learned shared and the private features are closely entangled, weakening the refining of the shared task-invariant fine-grained feature.
Therefore, we employ a third-party task discriminator with adversarial training to refine the features. 
The discriminator $D$ is a classifier for predicting what the current task is, based merely on the task-invariant fine-grained feature representation $\bm{v}^s$.
Ideally, once the discriminator cannot accurately identify the task ID $y^d_k$, the task-invariant fine-grained feature representation $\bm{v}^s$ can be understood as the most purified one. 
Specifically, the discriminator is a 4-layer 768-d Transformer (Trm) network, where we use a feedforward layer (FFN) with Softmax for the task prediction:
\begin{equation}
   \bm{v}^{'} = \text{Trm} (\bm{v}_1, \cdots, \bm{v}_n) ,
\end{equation}
\begin{equation}
    \bar{y}^d_k = \text{Softmax} (\text{FFN}(\bm{v}^{'})) ,
\end{equation}
where $\bar{y}^d_k$ is the predicted task ID.

The target of this adversarial training is to urge the shared features such that the discriminator cannot reliably predict the task ID:
\begin{equation}
    \mathcal{L}^{syn} = \text{min}_{\theta} (\text{max}_{D} ( \sum_k  \bar{y}^d_k \log (y^d_k) ) ).
\end{equation}

\subsection{Overall Training Remarks}

All our framework is trained through three main stages, in a specific ordering of sub-steps:

\begin{enumerate}
    \item[\bf $\bullet$ Step-1:] Basic Multimodal Comprehension and Generation Skill Training, cf $\S$\ref{sec:Basic Multimodal Comprehension and Generation Skill Training}.
    \begin{enumerate}
            \item[\bf $\bullet$ Step-1.1:] Aligning the encoder-LLM for overall vision-language alignment learning.
            
            \item[\bf $\bullet$ Step-1.2:] Doing text invocation instruction tuning such that the MLLM learns to output text instructions in the correct format.

            \item[\bf $\bullet$ Step-1.3:] When the above step is converging, training the LLM with continuous soft embedding-oriented LLM-decoder alignment, such that the LLM overall can convey the signal to the downstream modules.
    \end{enumerate}

    \item[\bf $\bullet$ Step-2:] Fine-grained Spatiotemporal Vision Grounding Instruction Tuning, cf $\S$\ref{Fine-grained Spatiotemporal Vision Grounding Instruction Tuning}.
    \begin{enumerate}
            \item[\bf $\bullet$ Step-2.1:] Starting by doing the Image Spatial Grounding training, on the grounded image captioning task and referring image segmentation task.
            
            \item[\bf $\bullet$ Step-2.2:] When MLLM has the ability for fine-grained spatial understanding, doing the Video Spatial-Temporal Grounding training, on the grounded video captioning task and referring video tracking task.

            \item[\bf $\bullet$ Step-2.3:] When the MLLM has learned to have the competent ability of both image and video spatiotemporal understanding at the perception level, doing the Grounding-aware Vision QA task at the cognition level.
    \end{enumerate}

    \item[\bf $\bullet$ Step-3:] As the final step, when the overall system has learned to have a competitive ability in various visual tasks, dining the cross-task synergy learning, cf $\S$\ref{Cross-task Synergy Learning}.
    This should be done by combining both the adversarial training ($\mathcal{L}^{syn}$) with the end-task prediction ($\mathcal{L}_{k}$).
    So the total loss of the step-3 is: $\mathcal{L}^{syn}+\sum_k\mathcal{L}_{k}$.

\end{enumerate}

\section{Extended Experimental Settings}
\label{Extended Details of Experimental Settings}

We quantify the performance of \textsc{Vitron} on a variety of standard benchmarks for downstream vision tasks and compare it against some of the currently strong-performing systems. 
Given the countless vision tasks within the community, our experiments focus only on 1-2 of the most representative tasks from each task category for validation.
To ensure a fair comparison, all subsequent experiments adopt settings same or similar to those of baseline systems, with evaluations following established practices. 
Before experiments, we perform targeted pre-training on all of \textsc{Vitron}'s backend modules (such as GLIGEN and SEEM) on their respective datasets. 
This ensures our system is optimized for the best possible performance during testing.
Our approach centers on training the linear projection layers of all encoders and efficiently fine-tuning the language model using LoRA.

Our backbone LLM is Vicuna\footnote{\url{https://huggingface.co/lmsys/vicuna-7b-v1.5}}, 7B, version 1.5.
The CLIP-ViT encoders for both images and videos are with a patch size of 14, and convert all images and video frames into 336px resolutions.
The task discriminator in our synergy module is with a Transformer architecture, with 4 layers and each in 768-d representation.
To train our model, we employ the AdamW optimizer along with a learning rate scheduler.
The pre-training of \textsc{Vitron} unfolds in three phases, all conducted on 10$\sim$16 $\times$ A100 (80G) GPUs. 
Initially, we train the model using a global batch size of 128 and a maximum learning rate of 3e-4, a process that takes approximately 40 hours. 
In the second tuning phase, we adjust the model with a maximum learning rate of 1e-5, utilizing a global batch size of 90. 
This stage of training lasts about 35 hours. 
The third phase of training employs a global batch size of 128 and maintains the maximum learning rate of 1e-5, completing in roughly 10 hours.

\section{More Experiment Results}
\label{More Experiment Results}

\subsection{Vision Segmentation}
\label{More Vision Segmentation}

\paragraph{Video Segmentation.}
Table \ref{tab:video-object-Segmentation-2} presents the comprehensive comparison of \textsc{Vitron} with some SoTA systems in video tracking on DAVIS 17 \cite{pont20172017} Test-Dev and Youtube-VOS 2019 \cite{xu2018youtube} Val sets.

\begin{table*}[!h]
\centering
\fontsize{8}{9}\selectfont
\setlength{\tabcolsep}{4.5mm}
\begin{tabular}{lccccccc}
\hline
\multirow{2}{*}{\bf  Method }& \multicolumn{3}{c}{\bf DAVIS 17 \cite{pont20172017} Test-Dev}&  \multicolumn{4}{c}{\bf Youtube-VOS 2019 \cite{xu2018youtube} Val} \\ 
\cmidrule(r){2-4} \cmidrule(r){5-8} 
& $\mathcal{J}$\&$\mathcal{F}$& 	$\mathcal{J}$& 	$\mathcal{F}$& 	$\mathcal{J}_s$& 	$\mathcal{F}_s$& 	$\mathcal{J}_u$	& $\mathcal{F}_u$\\

\hline
RDE \cite{li2022recurrent} &	77.4 &	73.6 &	81.2 &	81.1 &	85.5 &	 76.2  &	84.8 \\
XMem \cite{cheng2022xmem} &	81.0  &	77.4 &	84.5 &	84.3 &	 89.6 & 	 80.3 &	 88.6 \\
DeAOT \cite{yang2022decoupling} & 	 80.7 &	76.9  &	 84.5  &	 84.6 &	89.4 &	80.8 & 	88.9 \\
ISVOS \cite{wang2023look} &	82.8 &	79.3 &  	86.2 &	85.2 & 	89.7 &	 80.7 &	88.9 \\

\rowcolor{lightgreen} \textsc{Vitron} &	\bf84.2 &	\bf81.5 &	\bf86.7 &	\bf86.5 &	\bf90.4 &	\bf81.9 &	\bf90.2 \\

\hline
\end{tabular}
\caption{
Results of video object segmentation.
}
\vspace{-3mm}
\label{tab:video-object-Segmentation-2}
\end{table*}

\subsection{Fine-grained Vision Understanding}

\paragraph{Region-level Image Understanding.}

The comparisons of image-referring expression comprehension on three datasets are
shown in Tables \ref{tab:referring-expression-comprehension-2}.

\begin{table*}[!h]
\centering
\fontsize{8}{9}\selectfont
\setlength{\tabcolsep}{3.8mm}
\begin{tabular}{lcccccccc}
\hline
\multirow{2}{*}{\bf  Method }& \multicolumn{3}{c}{\bf RefCOCO \cite{kazemzadeh2014referitgame}}& \multicolumn{3}{c}{\bf RefCOCO+ \cite{yu2016modeling}} & \multicolumn{2}{c}{\bf RefCOCOg \cite{mao2016generation}} \\ 
\cmidrule(r){2-4} \cmidrule(r){5-7}  \cmidrule(r){8-9} 
& Val& 	TestA& 	TestB& 	Val& 	TestA& 	TestB& 	Val& 	Test \\
\hline
OFA \cite{wang2022ofa} &	 80.0 &	 83.7 &	 76.4 & 	 68.3 & 	 76.0  &	 61.8 &	 67.6 &	 67.6 \\
Shikra \cite{chen2023shikra} &	87.0 &	 90.6 &  	 80.2 &	81.6 & 	87.4 & 	 72.1 &	 82.3  &	 82.2 \\
MiniGPT-v2 \cite{chen2023minigpt} &	88.7 &	91.6 &	 85.3 &	79.9 &	 85.1 &	74.4 &	 84.4 &	 84.6 \\

\rowcolor{lightgreen} \textsc{Vitron}&	\bf 90.9 &	\bf 93.2 &	\bf 89.3 &	\bf 83.7 &	\bf 89.1 &	\bf 76.9 &	\bf 86.4 &	\bf 87.0 \\

\hline
\end{tabular}
\caption{
Results (accuracy) of image referring expression comprehension.
}
\label{tab:referring-expression-comprehension-2}
\end{table*}

Table \ref{tab:image-VQA-2} displays the results across 6 datasets for image-based VQA.

\begin{table*}[!h]
\centering
\fontsize{8}{9}\selectfont
\setlength{\tabcolsep}{1.2mm}
\begin{tabular}{lccccccc}
\hline

\bf Method & \bf Grounding& 	\bf OKVQA \cite{schwenk2022okvqa} & \bf GQA \cite{hudson2019gqa}& \bf  VSR \cite{liu2023visual}& \bf 	IconVQA \cite{lu2021iconqa}& 	\bf  VizWiz \cite{gurari2018vizwiz}	&\bf   HM \cite{kiela2020hateful} \\
\hline

Flamingo \cite{AlayracDLMBHLMM22} & 	\textcolor{ired}{\ding{55}}	 &    44.7 &   	  -  &  31.8  &  -  &  28.8  &  57.0 \\
BLIP-2 \cite{0008LSH23} & 	\textcolor{ired}{\ding{55}} & 	45.9 &  41.0  & 50.9 & 40.6  &  19.6  &   53.7 \\
InstructBLIP \cite{abs-2305-06500} &  	\textcolor{ired}{\ding{55}} & 	-  & 	49.5  &  52.1 & 44.8 &  33.4  & 57.5 \\
MiniGPT-4 \cite{abs-2304-10592} & 	\textcolor{ired}{\ding{55}} &  37.5  & 30.8	  & 41.6 & 	37.6  &    -   &   - \\
LLaVA \cite{abs-2304-08485} & 	\textcolor{ired}{\ding{55}} & 	54.4 &  41.3  & 51.2  & 43.0  &   -  &   - \\
Shikra \cite{chen2023shikra} & 	\textcolor{igreen}{\ding{51}} & 	 47.2 &  -  & 	 - &  	 - &     - &      - \\
MiniGPT-v2 \cite{chen2023minigpt} &  	\textcolor{igreen}{\ding{51}} & 	 57.8  & 	 60.1 &  	62.9 & 	51.5  & 	 53.6  & 	 58.8 \\

\rowcolor{lightgreen} \textsc{Vitron} & 	\textcolor{igreen}{\ding{51}} & \bf 59.4 & \bf 	62.1 & 	\bf 63.9 & 	\bf 52.2 & 	\bf 54.7 & 	\bf 60.2 \\

\hline
\end{tabular}
\caption{
Results (accuracy) on image-based VQA.
}
\vspace{-3mm}
\label{tab:image-VQA-2}
\end{table*}

\paragraph{Region-level Video Understanding.}

Table \ref{tab:video-QA-2} presents the results of video QA across 4 representative datasets. 
Interestingly, while PG-Video-LLaVA has video grounding capabilities, it does not show better results than Video-LLaVA, which lacks grounding. However, our \textsc{Vitron} achieves superior performance.

\begin{table*}[!h]
\centering
\fontsize{8}{9}\selectfont
\setlength{\tabcolsep}{1mm}
\begin{tabular}{lccccccccc}
\hline
\multirow{2}{*}{\bf  Method }&\multirow{2}{*}{\bf  Grounding }& \multicolumn{2}{c}{\bf MSVD-QA \cite{xu2017video}}& \multicolumn{2}{c}{\bf MSRVTT-QA \cite{XuMYR16}} & \multicolumn{2}{c}{\bf TGIF-QA \cite{li2016tgif}}  & \multicolumn{2}{c}{\bf ActivityNet-QA \cite{yu2019activitynet}} \\ 
\cmidrule(r){3-4} \cmidrule(r){5-6}  \cmidrule(r){7-8}  \cmidrule(r){9-10} 
 & & \bf Accuracy& \bf Score& \bf Accuracy& \bf Score& \bf Accuracy& \bf Score& \bf Accuracy& \bf Score \\
\hline
VideoChat \cite{abs-2305-06355} &	\textcolor{ired}{\ding{55}} &	 56.3 & 	2.8	 &   45.0  & 	2.5	  &  34.4 &   	2.3	 &    - &	2.2 \\
LLaMA-Adapter \cite{gao2023llama} &	\textcolor{ired}{\ding{55}} &	 54.9 & 	3.1 &	  43.8 &  	2.7 &	54.3 &	3.3	   & 34.2 &	2.7 \\
Video-LLaMA \cite{abs-2306-02858}  &	\textcolor{ired}{\ding{55}} &	51.6  &	2.5	 &   29.6 &  	1.8 &	51.4  &	3.4	   &12.4	 &1.1 \\
Video-ChatGPT \cite{abs-2306-05424} &	\textcolor{ired}{\ding{55}} &	 64.9  &	3.3 &	  49.3 &  	2.8 &	 51.4  &	3.0	  & 35.2 &	2.7 \\
Video-LLaVA \cite{lin2023video}  &	\textcolor{ired}{\ding{55}}	 &  70.7 &	3.9	 &  59.2 &	3.5	 & 70.0 &	4.0	 &    45.3 &	3.3 \\
PG-Video-LLaVA \cite{munasinghe2023pg} &	\textcolor{igreen}{\ding{51}} &	64.1 &	3.7 &	51.6  &	3.3  &	66.8 & 	3.8  &	39.9  &	3.3 \\

\rowcolor{lightgreen} \textsc{Vitron} &	\textcolor{igreen}{\ding{51}} &		\bf74.9 &		\bf4.0 &		\bf62.0 &		\bf3.8 &		\bf72.4 &		\bf4.1 &		\bf51.0 &	\bf	3.7 \\

\hline
\end{tabular}
\caption{
Results (accuracy and confidence Score) on video QA.
}
\vspace{-3mm}
\label{tab:video-QA-2}
\end{table*}

\vspace{-3mm}

\section{Qualitative Studies of Case Visualizations}

\vspace{-2mm}
\subsection{Vision Segmentation}

Fig. \ref{fig:image-Segmentation} further demonstrates an example of how our \textsc{Vitron} processes image segmentation tasks in an interactive manner with the user.
When users sketch or doodle outlines on specific areas of an image, \textsc{Vitron} is capable of accurately identifying the corresponding objects within the image. 
Following this, it precisely generates the bounding box and mask area for the identified objects.

Fig. \ref{fig:video-Segmentation} also demonstrates the process of video segmentation.
\textsc{Vitron} successfully identifies, localizes, and tracks the target in the video based on the provided queries.
Our system demonstrates a remarkable ability to accurately and flexibly capture the user's intent, even when the desired object to be segmented is described in a highly implicit manner.

\begin{minipage}{0.49\textwidth} 
\centering
\includegraphics[width=1\columnwidth]{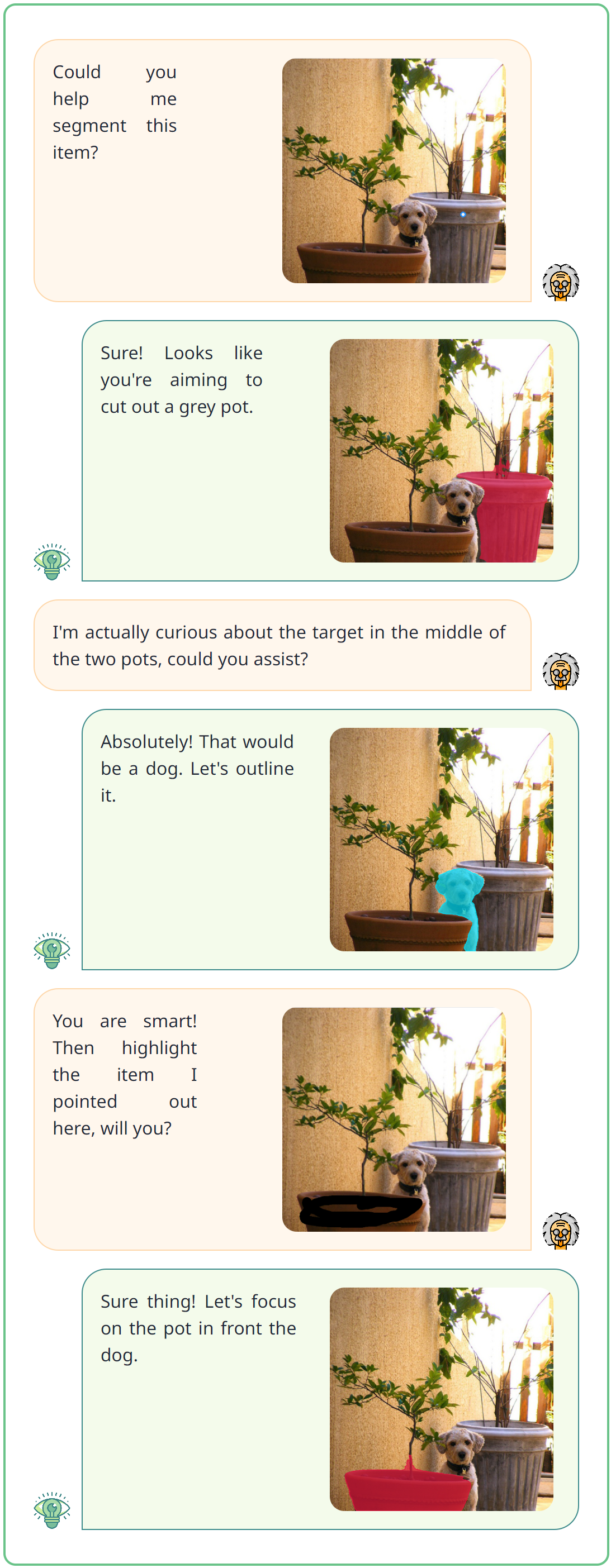}
\captionof{figure}{
Demonstration of image segmentation.
}
\label{fig:image-Segmentation}
\end{minipage}%
\hfill
\begin{minipage}{0.49\textwidth} 
\centering
\includegraphics[width=1\columnwidth]{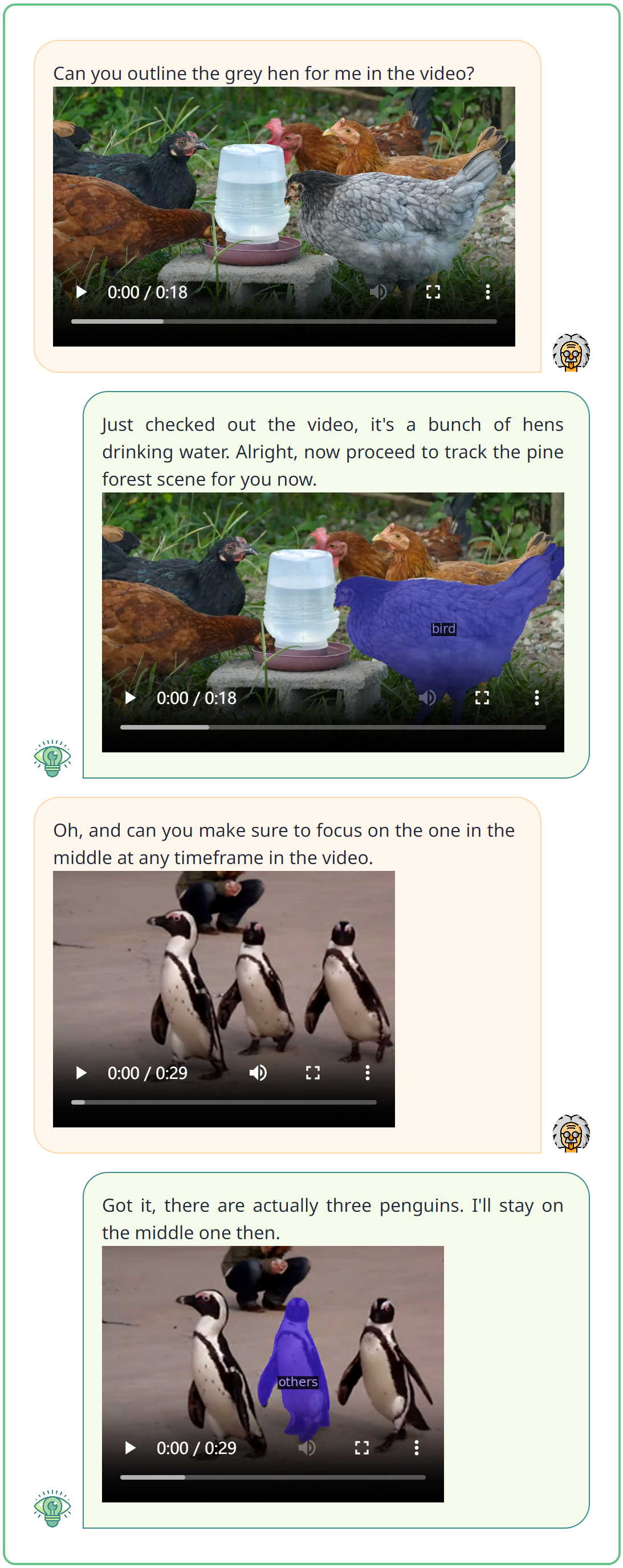}
\captionof{figure}{
Example of video object segmentation.
}
\label{fig:video-Segmentation}
\end{minipage}

\subsection{Fine-grained Vision Understanding}

Fig. \ref{fig:image-grounding} highlights \textsc{Vitron}'s fine-grained visual understanding capabilities through regional image captioning, to accurately understand and appropriately describe image regions.

Further, we showcase a visualization process illustrating how \textsc{Vitron} accomplishes fine-grained visual understanding through video grounding tasks in Fig. \ref{fig:video-grounding}.
\textsc{Vitron}'s precise perception of video details ensures a more accurate understanding of content semantics, leading to stronger video understanding capabilities.

\begin{minipage}{0.49\textwidth} 
\centering
\includegraphics[width=1\columnwidth]{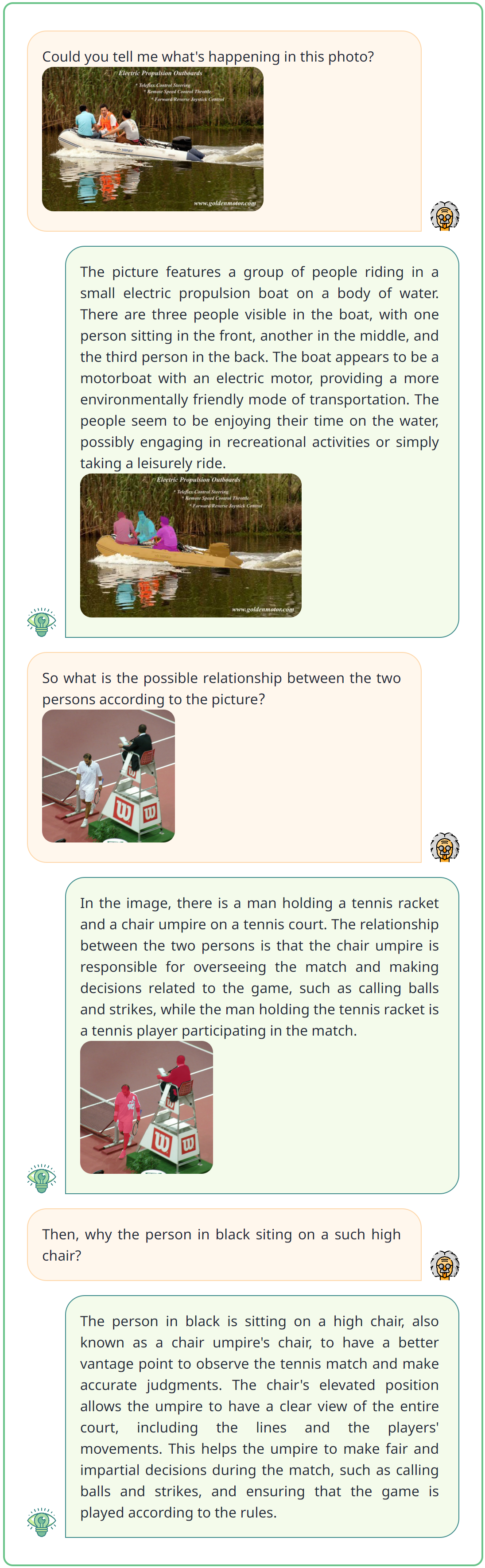}
\captionof{figure}{
Demonstration of fine-grained visual understanding via image grounding.
}
\label{fig:image-grounding}
\end{minipage}%
\hfill
\begin{minipage}{0.49\textwidth} 

\begin{minipage}{\textwidth}
\centering
\includegraphics[width=1\columnwidth]{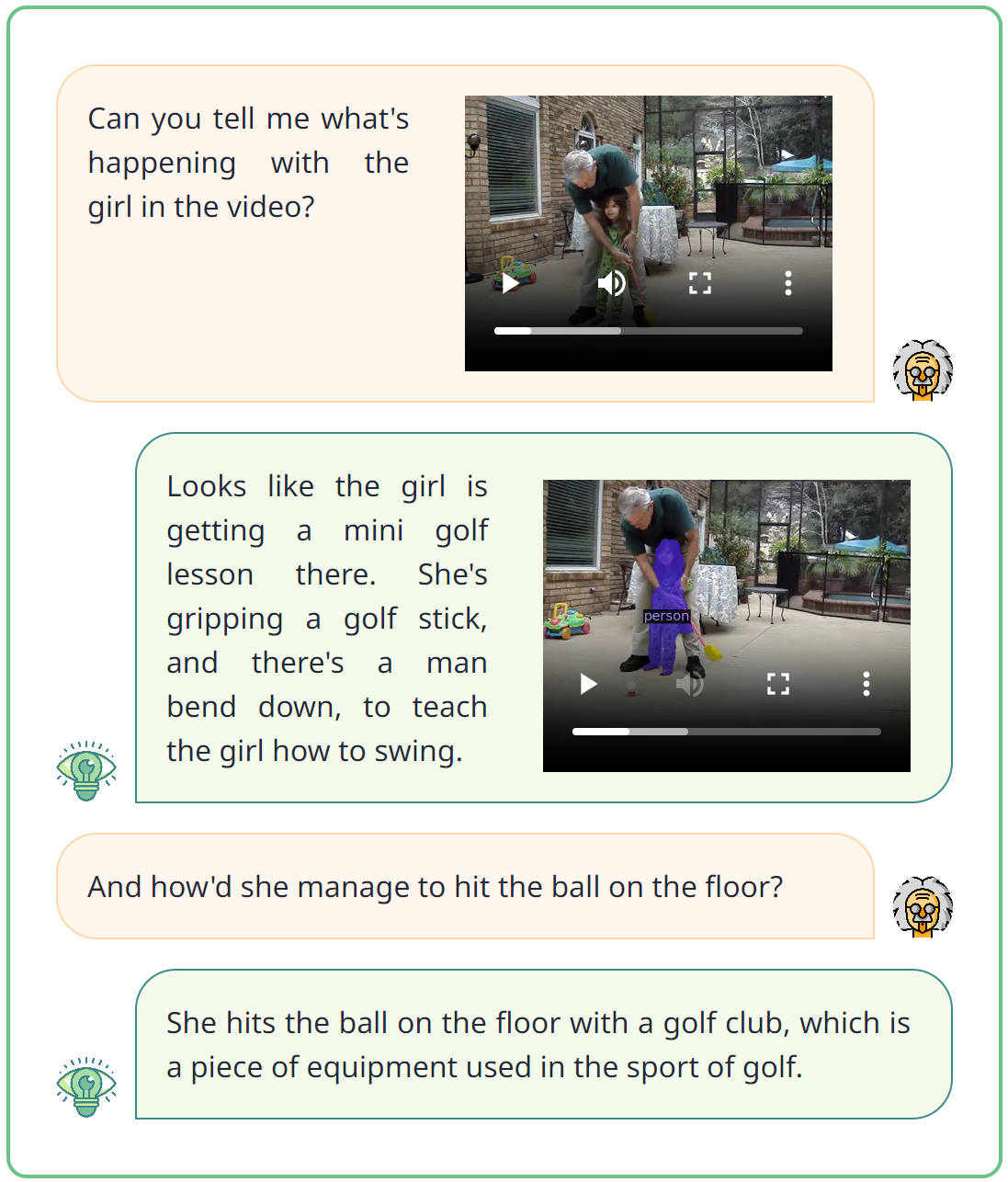}
\captionof{figure}{
Demonstration of fine-grained visual understanding via video grounding.
}
\label{fig:video-grounding}
\end{minipage}
\vspace{4pt} 

\begin{minipage}{\textwidth}
\centering
\includegraphics[width=1\columnwidth]{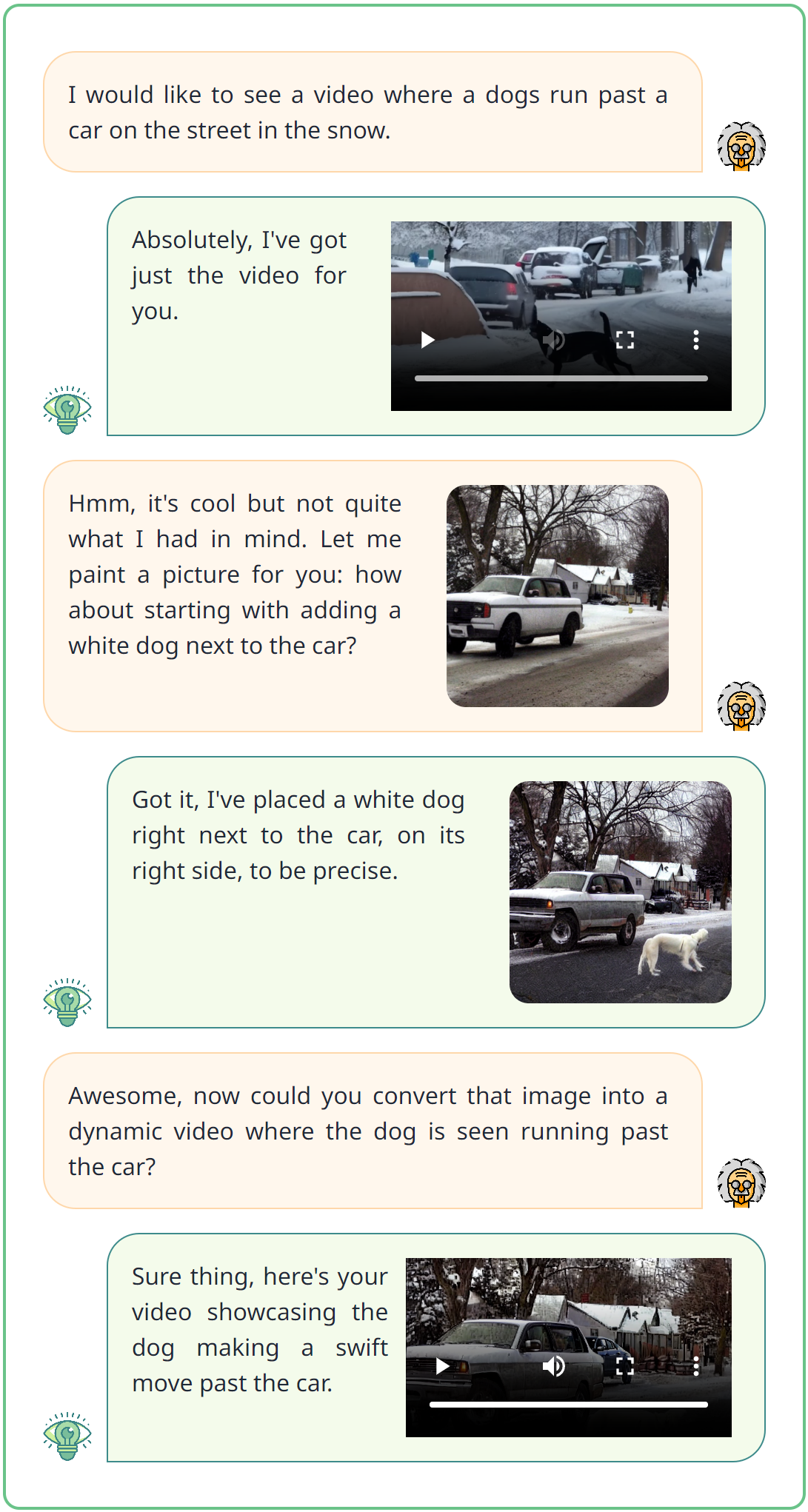}
\captionof{figure}{
Demonstration of vision generation across different modalities.
}
\label{fig:vision-generation}
\end{minipage}

\end{minipage}

\subsection{Vision Generation}

Fig. \ref{fig:vision-generation} illustrates the process of vision generation across different modalities, including text, image, and video. 
Initially, users start with a basic text command, and \textsc{Vitron} is capable of transforming a simple idea into a detailed video. 
However, if users are not satisfied with the video generated directly from text, they can first generate an image from text, then fine-tune or edit this image, and finally create a satisfying video based on the adjusted image.
Our \textsc{Vitron}, thanks to its robust interactive capability via multi-turn dialogue, enables users to perform a series of consecutive operations, ultimately facilitating smooth content creation. 
This fully helps meet the demands of real-world application scenarios.

\subsection{Vision Editing}

\begin{minipage}{0.49\textwidth} 
\centering
\includegraphics[width=1\columnwidth]{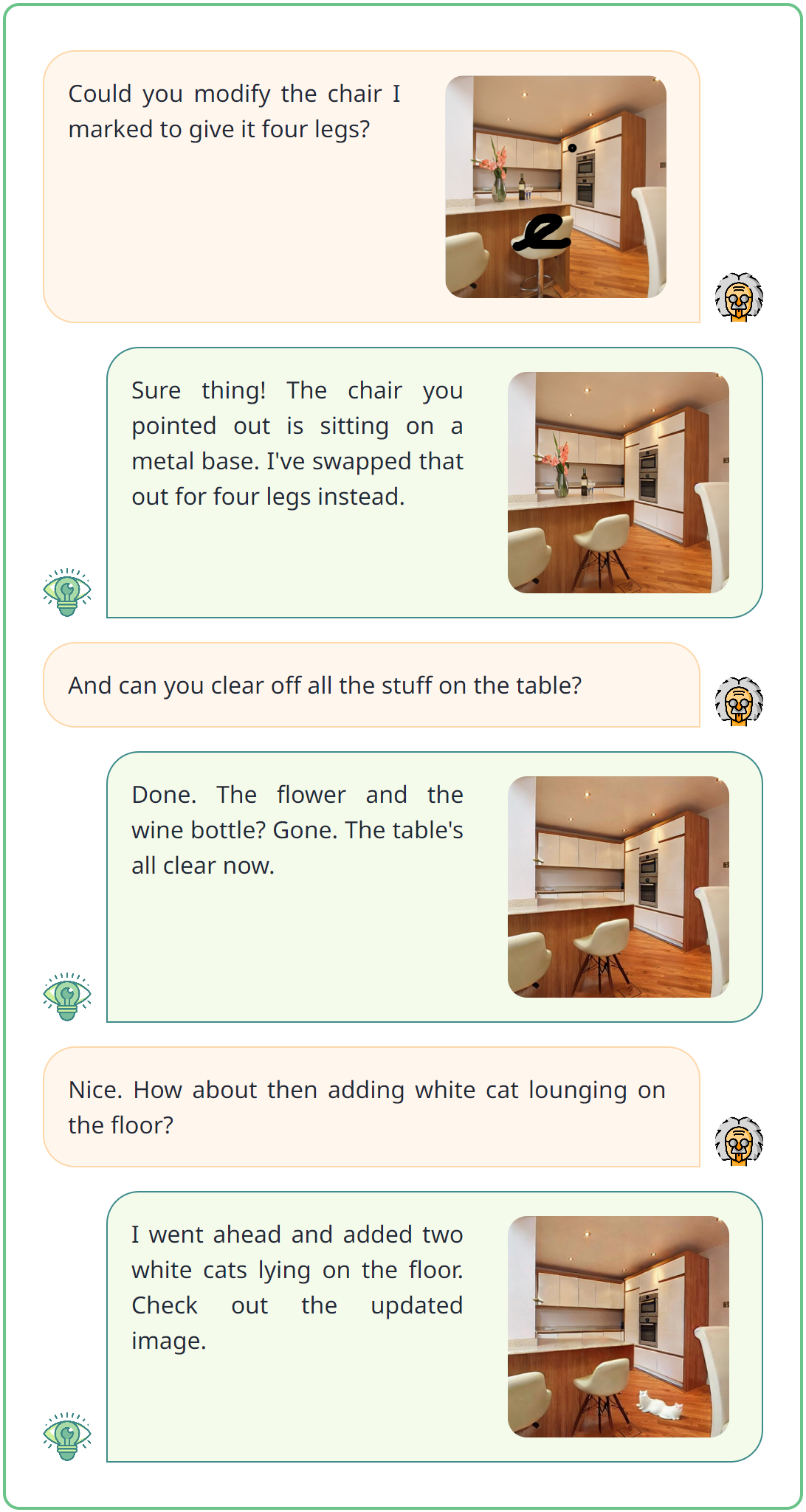}
\captionof{figure}{
Demonstration of image editing.
}
\label{fig:image-editing}
\end{minipage}%
\hfill
\begin{minipage}{0.49\textwidth} 
\centering
\includegraphics[width=1\columnwidth]{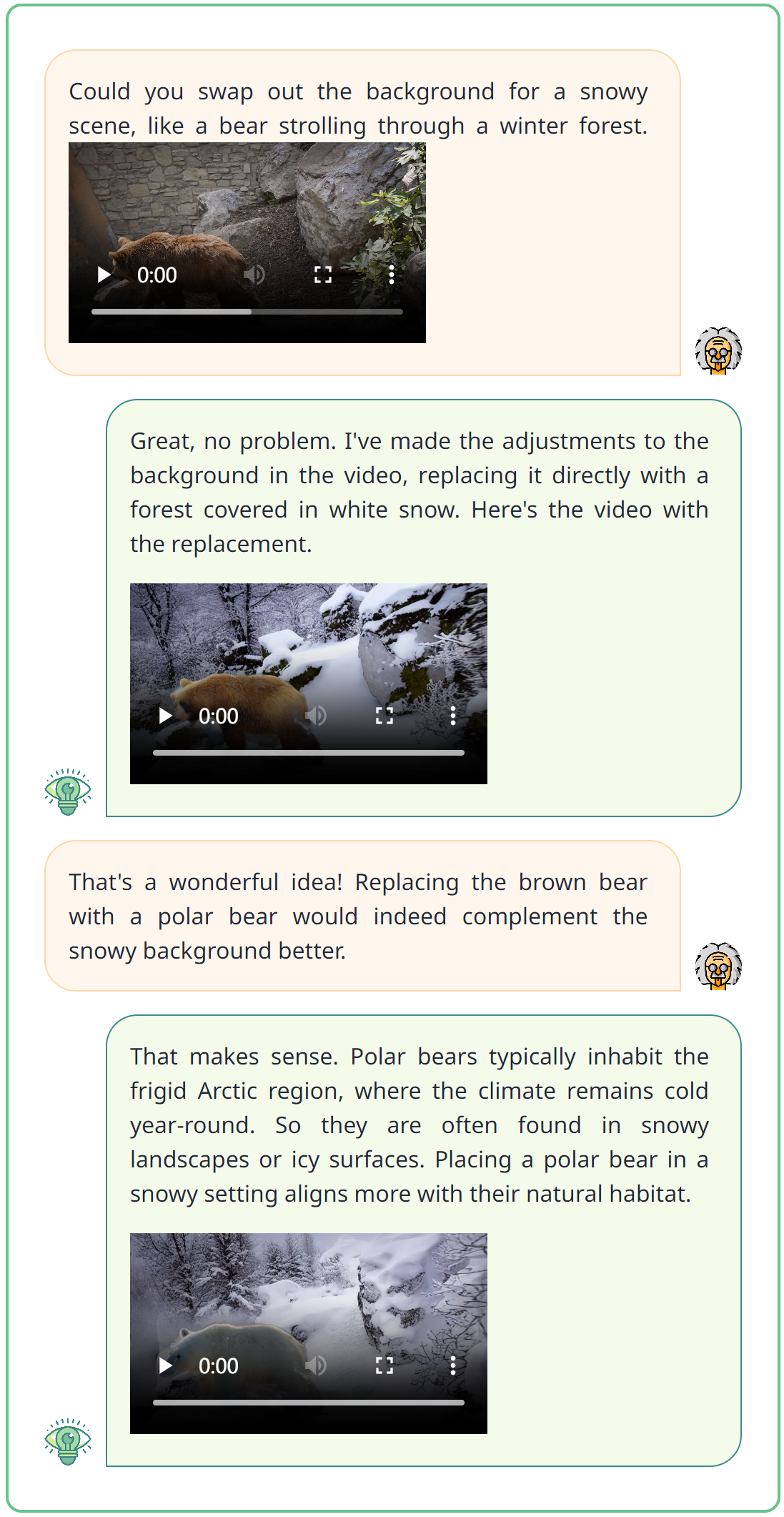}
\captionof{figure}{
Demonstration of video editing.
}
\label{fig:video-editing}
\end{minipage}

We showcase the specific process of this image editing, as illustrated in Fig. \ref{fig:image-editing}. 
\textsc{Vitron} is capable of accepting different forms of user inputs (textual instruction or sketch) for precise image edits. 
It maintains contextual consistency throughout a series of sequential editing operations, ultimately achieving satisfactory results that meet the user's expectations.

Fig. \ref{fig:video-editing} illustrates this process. \textsc{Vitron} competently handles video editing tasks, including modifications to the content's subject, and changes to the video's style, etc.


\newpage
\section*{NeurIPS Paper Checklist}

\begin{enumerate}

\item {\bf Claims}
    \item[] Question: Do the main claims made in the abstract and introduction accurately reflect the paper's contributions and scope?
    \item[] Answer: \answerYes{} 
    \item[] Justification: in the Section \ref{Introduction} \& \ref{related_work}

\item {\bf Limitations}
    \item[] Question: Does the paper discuss the limitations of the work performed by the authors?
    \item[] Answer: \answerYes{} 
    \item[] Justification: in the Experiment part.

\item {\bf Theory Assumptions and Proofs}
    \item[] Question: For each theoretical result, does the paper provide the full set of assumptions and a complete (and correct) proof?
    \item[] Answer: \answerYes{} 
    \item[] Justification: in the Section \ref{architecture} \& \ref{sec:Pixel-aware Synergistic Vision-Language Understanding Tuning}   

    \item {\bf Experimental Result Reproducibility}
    \item[] Question: Does the paper fully disclose all the information needed to reproduce the main experimental results of the paper to the extent that it affects the main claims and/or conclusions of the paper (regardless of whether the code and data are provided or not)?
    \item[] Answer: \answerYes{} 
    \item[] Justification: in the Appendix \ref{details_of_visual_backbone} and Appendix \ref{Extended Details of Experimental Settings}

\item {\bf Open access to data and code}
    \item[] Question: Does the paper provide open access to the data and code, with sufficient instructions to faithfully reproduce the main experimental results, as described in supplemental material?
    \item[] Answer: \answerYes{} 
    \item[] Justification: All the data utilized in our experiments are publicly available. We will release the code upon the acceptance of the paper.

\item {\bf Experimental Setting/Details}
    \item[] Question: Does the paper specify all the training and test details (e.g., data splits, hyperparameters, how they were chosen, type of optimizer, etc.) necessary to understand the results?
    \item[] Answer: \answerYes{} 
    \item[] Justification: in the Appendix \ref{Extended Details of Experimental Settings}

\item {\bf Experiment Statistical Significance}
    \item[] Question: Does the paper report error bars suitably and correctly defined or other appropriate information about the statistical significance of the experiments?
    \item[] Answer: \answerYes{} 
    \item[] Justification: in the Section \ref{experiments} and Appendix \ref{More Experiment Results}, and All results are reported after the statistical significance tests.

\item {\bf Experiments Compute Resources}
    \item[] Question: For each experiment, does the paper provide sufficient information on the computer resources (type of compute workers, memory, time of execution) needed to reproduce the experiments?
    \item[] Answer: \answerYes{} 
    \item[] Justification: in the Section \ref{Extended Details of Experimental Settings}
    
\item {\bf Code Of Ethics}
    \item[] Question: Does the research conducted in the paper conform, in every respect, with the NeurIPS Code of Ethics \url{https://neurips.cc/public/EthicsGuidelines}?
    \item[] Answer: \answerYes{} 
    \item[] Justification: the research conducted in the paper conforms, in every respect, with the NeurIPS Code of Ethics 

\item {\bf Broader Impacts}
    \item[] Question: Does the paper discuss both potential positive societal impacts and negative societal impacts of the work performed?
    \item[] Answer: \answerNo{} 
    \item[] Justification: \answerNA{}
    
\item {\bf Safeguards}
    \item[] Question: Does the paper describe safeguards that have been put in place for responsible release of data or models that have a high risk for misuse (e.g., pretrained language models, image generators, or scraped datasets)?
    \item[] Answer: \answerNA{} 
    \item[] Justification: the paper poses no such risks.

\item {\bf Licenses for existing assets}
    \item[] Question: Are the creators or original owners of assets (e.g., code, data, models), used in the paper, properly credited and are the license and terms of use explicitly mentioned and properly respected?
    \item[] Answer: \answerNA{} 
    \item[] Justification: the paper does not use existing assets.

\item {\bf New Assets}
    \item[] Question: Are new assets introduced in the paper well documented and is the documentation provided alongside the assets?
    \item[] Answer: \answerNA{} 
    \item[] Justification: the paper does not release new assets.

\item {\bf Crowdsourcing and Research with Human Subjects}
    \item[] Question: For crowdsourcing experiments and research with human subjects, does the paper include the full text of instructions given to participants and screenshots, if applicable, as well as details about compensation (if any)? 
    \item[] Answer: \answerNA{} 
    \item[] Justification: the paper does not involve crowdsourcing nor research with human subjects.

\item {\bf Institutional Review Board (IRB) Approvals or Equivalent for Research with Human Subjects}
    \item[] Question: Does the paper describe potential risks incurred by study participants, whether such risks were disclosed to the subjects, and whether Institutional Review Board (IRB) approvals (or an equivalent approval/review based on the requirements of your country or institution) were obtained?
    \item[] Answer: \answerNA{} 
    \item[] Justification:  the paper does not involve crowdsourcing nor research with human subjects.

\end{enumerate}

\end{document}